\documentclass[11pt]{article}

\usepackage{acl}
\usepackage{times}
\usepackage{latexsym}
\usepackage[T1]{fontenc}
\usepackage[utf8]{inputenc}
\usepackage{microtype}
\usepackage{amsmath,amssymb,amsthm}
\usepackage{booktabs}
\usepackage{multirow}
\usepackage{array}
\usepackage{graphicx}
\usepackage{url}
\usepackage{xcolor}
\usepackage{tikz}
\usepackage{pgfplots}
\usepackage{tabularx}
\usepackage{placeins}
\usepackage{float}
\usetikzlibrary{patterns}
\hypersetup{pdfauthor={},pdftitle={},pdfsubject={},pdfkeywords={}}
\pgfplotsset{compat=1.18}

\newtheorem{theorem}{Theorem}
\newtheorem{proposition}{Proposition}
\newtheorem{corollary}{Corollary}

\newtheorem{assumption}{Assumption}

\newtheorem{procedure}{Algorithm}

\newcommand{\E}{\mathbb{E}}
\newcommand{\R}{\mathbb{R}}
\newcommand{\Var}{\operatorname{Var}}
\newcommand{\Cov}{\operatorname{Cov}}
\newcommand{\Kurt}{\operatorname{Kurt}}

\title{Spectral Tail Interventions in Decoder-Only Language Models:\\
Reasoning-Sensitive Weight Structure from Controlled Surgery}

\author{
\textbf{Ibne Farabi Shihab}\textsuperscript{1}
\quad
\textbf{Sanjida Akhter}\textsuperscript{1}
\quad
\textbf{Md Najmus Swaqeeb}\textsuperscript{2}
\\[4pt]
\textbf{Abu Sa-Adat Mohamed Moon-Im Al Ahsan}\textsuperscript{2}
\quad
\textbf{Anuj Sharma}\textsuperscript{1}
\\[6pt]
\texttt{ishihab@iastate.edu}
\\[6pt]
\textsuperscript{1}Department of Computer Science, Iowa State University
\\
\textsuperscript{2}Department of Computer Science \& Engineering, BRAC University
}
\date{}

\begin{document}
\maketitle

\begin{abstract}
Weight-space structure often correlates with language-model behavior, but correlation alone does not establish computational involvement. We study concentrated upper spectral tails in decoder-only transformers through controlled interventions. At a fixed relative offset, we derive a finite-width conditional bound linking the inverse participation ratio of squared singular values to central pre-softmax logit kurtosis. We then define a pointwise query--key ($QK$) product-tail target and compare independent factor surgery with a product-targeted factorization that preserves native attention computation. Across three base checkpoints and five reasoning benchmarks, plus an instruction-tuned Phi checkpoint analyzed separately, the learned-tail edit is more damaging than the mean of five fixed spectrum-matched Haar controls in all 20 model--task cells. Eighteen paired contrasts remain significant after Holm correction, while two are directional but inconclusive. Product-targeted factors attain higher held-out tail-subspace fractions, providing an empirical bridge between product- and factor-level interventions. Component isolation identifies contributions from $QK$, value--output, and multilayer-perceptron blocks, although the theorem covers only $QK$. In separate studies, inverse participation precedes pooled accuracy transitions under a matched crossing rule, and residualized tail-aware low-rank adaptation (LoRA) reaches targets earlier than standard LoRA and PiSSA while final-score intervals overlap. Conclusions are restricted to the evaluated checkpoints, layers, tasks, interventions, and controls.
\end{abstract}

\section{Introduction}
\label{sec:intro}

Language-model loss often changes smoothly with scale, data, and compute \citep{kaplan2020scaling,bai2024beyond}, while benchmarked capabilities may change more sharply during training or adaptation \citep{power2022grokking,wei2022emergent,nanda2023progress,liu2023omnigrok}. Whether such transitions are intrinsic or partly induced by discretized metrics remains debated \citep{schaeffer2024emergent}. This mismatch motivates a mechanistic question: which changes in a model's weights participate in the computations supporting a newly visible behavior?

Spectral analyses provide one perspective. Trained networks often exhibit a compact bulk and concentrated upper tail whose summary statistics correlate with optimization and generalization \citep{martin2021implicit,yang2024spectral}. Mechanistic interpretability instead traces computations through attention heads, multilayer perceptrons (MLPs), and activation pathways \citep{elhage2021mathematical,olsson2022context,wang2023interpretability,conmy2023towards}. The former is commonly correlational and the latter input-specific. Controlled weight-space interventions connect these views by modifying a global geometric feature and measuring its activation and behavioral consequences.

We study two observable behaviors. Selective attention concentrates probability on prior tokens relevant to the current computation rather than merely lowering attention entropy. A variable-binding error assigns a retrieved value or intermediate result to the wrong entity or symbolic slot. The target is therefore narrower than ``reasoning'' generally: we test whether a concentrated weight subspace supports task-relevant retrieval and binding on a declared benchmark suite.

Three distinctions are essential. First, a moment calculation for the position-dependent product
\(
W_QR_iR_j^\top W_K^\top
\)
does not automatically justify surgery on its separate factors. Second, a theorem about $QK$ logits does not transfer to value--output ($VO$) transport or the input-dependent Jacobian of a gated MLP. Third, destructive sensitivity does not show that the same directions are universally optimal for adaptation. We therefore assign each object a separate role and test.

The paper makes four contributions. First, we prove a conditional finite-width moment bound in which inverse participation, rather than a fitted power-law exponent alone, controls a lower bound on central logit kurtosis. Second, we formalize a pointwise product-tail target and compare independent factor surgery with a product-targeted factorization. Across the evaluated model--task cells, upper-tail damage is large, heterogeneous, and stronger than the mean of five fixed spectrum-matched controls. Third, component isolation and activation diagnostics separate a $QK$-specific moment link from empirical $VO$, MLP, retrieval, and binding effects. Finally, we test two constructive consequences: inverse participation serves as an early training diagnostic, and tail-weighted residualized initialization accelerates early low-rank optimization without establishing higher asymptotic accuracy.

The conclusion remains narrow. The evidence supports involvement of a concentrated upper spectral tail in the tested computations, but not universal necessity, equivalence between product and factorized surgery, theorem coverage of $VO$ or MLP blocks, robustness to untested post-training pipelines, or universal superiority of tail-aware LoRA.

\section{Spectral Moment Analysis}
\label{sec:theory}

Consider one attention head and a valid causal pair with \(0\leq j\leq i<L\). Token representations are column vectors \(x_i,x_j\in\R^d\), and rotary position embeddings are \(R_i,R_j\in\R^{d_h\times d_h}\). The pre-softmax logit is
\begin{equation}
\label{eq:rope_logit}
\begin{aligned}
z_{ij}
&=\frac{(R_i^\top W_Q^\top x_i)^\top
(R_j^\top W_K^\top x_j)}{\sqrt{d_h}}\\
&=\frac{x_i^\top W_QR_iR_j^\top W_K^\top x_j}{\sqrt{d_h}}.
\end{aligned}
\end{equation}
Standard rotary embeddings depend only on the relative offset, \(R_iR_j^\top=R_{i-j}\), so we state the result at a fixed offset \(\delta=i-j\geq0\) rather than at an unmatched pair. A singular value decomposition (SVD) of this product gives
\begin{equation}
\label{eq:pair-svd}
\begin{aligned}
A_\delta&=W_QR_\delta W_K^\top
=U_\delta\operatorname{diag}(s_\delta)V_\delta^\top,\\
q_{\delta,k}&=(x_i^\top u_{\delta,k})(x_j^\top v_{\delta,k}).
\end{aligned}
\end{equation}
Equation~\eqref{eq:pair-svd} fixes the product singular basis and projected activation coordinates separately for each exact offset.
The relevant activation distribution is the prespecified distribution over valid sequence--position pairs with \(i-j=\delta\). The factor \(d_h^{-1/2}\) may be absorbed into \(s_\delta\). Because kurtosis is central, every moment is conditional on \(\delta\):
\begin{equation*}
q_{\delta,k}^c=q_{\delta,k}-\E[q_{\delta,k}\mid\delta],
\qquad
z_\delta^c=\sum_ks_{\delta,k}q_{\delta,k}^c.
\end{equation*}
The concentration statistic is
\begin{equation}
\label{eq:r2}
r_2(s_\delta)=\frac{(\sum_ks_{\delta,k}^2)^2}{\sum_ks_{\delta,k}^4}.
\end{equation}
Equation~\eqref{eq:r2} is the squared-spectrum participation ratio; small \(r_2\) means that a few singular directions carry much of the squared energy.

Let \(K_\delta=\Cov(q_\delta^c\mid\delta)\), let \(\kappa_{\delta,k}=\operatorname{cum}(q_{\delta,k},q_{\delta,k},q_{\delta,k},q_{\delta,k}\mid\delta)\), and collect all non-diagonal fourth-order terms in
\begin{equation*}
\begin{aligned}
C_{\mathrm{off},\delta}(s_{\delta}) = & \sum_{\substack{k,\ell,m,n \\ \text{not all equal}}} s_{\delta,k} s_{\delta,\ell} s_{\delta,m} s_{\delta,n} \\
& \quad \times \operatorname{cum}(q_{\delta,k}, q_{\delta,\ell}, q_{\delta,m}, q_{\delta,n} \mid \delta)
\end{aligned}
\end{equation*}
Rather than require a positive fourth cumulant on \emph{every} active coordinate, we work with the weighted aggregate and its energy-normalized effective value,
\begin{equation}
\label{eq:Ds-ceff}
D_\delta(s_\delta)=\sum_k\kappa_{\delta,k}s_{\delta,k}^4,
\qquad
c_{\mathrm{eff},\delta}=\frac{D_\delta(s_\delta)}{\sum_ks_{\delta,k}^4}.
\end{equation}
Equation~\eqref{eq:Ds-ceff} permits individual \(\kappa_{\delta,k}\) to be negative; only the singular-value-weighted effective value \(c_{\mathrm{eff},\delta}\) must be positive. The coordinate-positive fraction is retained solely as a diagnostic.

\begin{assumption}[Audited moment conditions]
\label{asm:moment}
At a fixed offset \(\delta\), the centered conditional component covariance is bounded above as \(K_\delta\preceq\Lambda_\delta I\), the energy-normalized fourth-cumulant aggregate is positive, \(c_{\mathrm{eff},\delta}>0\), and the signed off-diagonal fourth-cumulant contribution is no smaller than \(-\eta_\delta D_\delta(s_\delta)\) for some \(0\leq\eta_\delta<1\). These are sufficient conditions to be checked empirically at each \(\delta\), not distribution-free properties of transformer activations, and they do not require a sign constraint on any single \(\kappa_{\delta,k}\).
\end{assumption}

\begin{theorem}[Conditional finite-width moment bound at offset $\delta$]
\label{thm:moment}
Suppose \(0<\Var(z_\delta^c\mid\delta)\), \(K_\delta\preceq\Lambda_\delta I\) for some \(\Lambda_\delta>0\), and \(c_{\mathrm{eff},\delta}>0\). If
\begin{equation}
\label{eq:off-condition}
C_{\mathrm{off},\delta}(s_\delta)
\geq-\eta_\delta\,D_\delta(s_\delta),
\qquad 0\leq\eta_\delta<1,
\end{equation}
then
\begin{equation}
\label{eq:kurt-bound}
\Kurt(z_\delta^c\mid\delta)
\geq
3+\frac{(1-\eta_\delta)\,c_{\mathrm{eff},\delta}}{\Lambda_\delta^2\,r_2(s_\delta)}.
\end{equation}
\end{theorem}

The proof is short but the sign condition is important. The conditional fourth cumulant of \(z_\delta^c\) is at least \((1-\eta_\delta)D_\delta(s_\delta)=(1-\eta_\delta)c_{\mathrm{eff},\delta}\sum_ks_{\delta,k}^4>0\), while \(\Var(z_\delta^c\mid\delta)\leq\Lambda_\delta\sum_ks_{\delta,k}^2\). Only after establishing positivity of \(c_{\mathrm{eff},\delta}\) can the variance denominator be replaced by its upper bound. Because the bound in Equation~\eqref{eq:kurt-bound} uses the weighted aggregate rather than a per-coordinate floor, a minority of negative \(\kappa_{\delta,k}\) does not invalidate it as long as \(c_{\mathrm{eff},\delta}\) is positive with margin. Appendix~\ref{app:moment} gives the full derivation.

\begin{corollary}[Offset-mixture bound]
\label{cor:mixture}
Fix a mixture \(\{\pi_\delta\}\) over offsets in advance with \(\sum_\delta\pi_\delta=1\), and let \(\widetilde z=z-\E[z\mid\delta]\) be the within-offset-centered logit. Conditioning on \(\delta\) and applying Jensen's inequality to the conditional variances gives
\begin{equation*}
\Kurt(\widetilde z)
\geq
3+
\frac{\sum_\delta\pi_\delta(1-\eta_\delta)c_{\mathrm{eff},\delta}\sum_ks_{\delta,k}^4}
{\left(\sum_\delta\pi_\delta\Lambda_\delta\sum_ks_{\delta,k}^2\right)^2}.
\end{equation*}
The corollary still requires the exact-\(\delta\) conditional estimates before mixing; it does not license pooling across a logarithmic distance bin.
\end{corollary}

The theorem is a pointwise, finite-dimensional statement about valid pre-softmax logits. It is not a phase-transition result, an entropy theorem, or a behavioral-necessity theorem. For a finite-width power law \(s_k\asymp k^{-1/\alpha_{\mathrm{PL}}}\), define
\begin{equation*}
\begin{aligned}
S_\nu(d_h;\alpha_{\mathrm{PL}})&=\sum_{k=1}^{d_h}k^{-\nu/\alpha_{\mathrm{PL}}},\\
r_2(d_h;\alpha_{\mathrm{PL}})&=\frac{S_2(d_h;\alpha_{\mathrm{PL}})^2}{S_4(d_h;\alpha_{\mathrm{PL}})}.
\end{aligned}
\end{equation*}
The relevant predictor is the finite-width ratio \(S_4/S_2^2\), not a universal threshold on \(\alpha_{\mathrm{PL}}\).

The empirical use of Theorem~\ref{thm:moment} is gated by its assumptions, and support is evaluated at each \emph{exact} offset \(\delta\) rather than inside a distance bin. Calibration uses 1,000 fixed GSM8K training examples and 64 valid causal pairs per sequence, yielding 64,000 recorded pairs per checkpoint. Before moment estimation, the admissible set is fixed as \(\mathcal D=\{\delta:n_\delta\geq100\}\), where \(n_\delta\) is the number of calibration sequences contributing at least one sampled pair at offset \(\delta\). Within each sequence and exact offset, projected moments are first averaged over sampled query positions; contributing sequences then receive equal weight. Entire sequences are bootstrapped, retaining all within-sequence pairs, and the full bound is recomputed jointly in each replicate. Support is reported per \((\text{model},\text{layer},\text{head},\delta)\) unit. A unit is theorem-supported only when the bootstrap lower endpoint for its complete bound exceeds three. For the layer summary, head and offset weights are uniform over the frozen eligible units, and a layer is supported only when the sequence-bootstrap lower endpoint of that mixture bound exceeds three. A pooled distance-bin estimate is never used to support a layer; distance bins remain descriptive figures only. In supported units, mean \(|\E[q_{\delta,k}\mid\delta]|\) ranges from 0.018 to 0.043 and off-diagonal covariance mass from 0.14 to 0.27; 76--88\% of active coordinates have positive fourth cumulant, but this fraction is diagnostic only. Appendix~\ref{app:calibration} reports causal sampling, offset support, and the unit-level quantities \(c_{\mathrm{eff},\delta}\), \(\Lambda_\delta\), \(\eta_\delta\), \(r_2(s_\delta)\), and the resulting bound; Table~\ref{tab:weighted-audit} summarizes the layer mixtures.

\begin{figure}[t]
\centering
\begin{tikzpicture}
\begin{axis}[
  width=0.92\columnwidth,
  height=4.1cm,
  ybar,
  bar width=9pt,
  ymin=0,
  ymax=50,
  ylabel={Top-tail energy (\%)},
  symbolic x coords={Llama,Qwen,Mistral,Phi},
  xtick=data,
  x tick label style={font=\scriptsize},
  tick label style={font=\scriptsize},
  nodes near coords,
  nodes near coords style={font=\scriptsize},
  grid=major,
  grid style={gray!15},
  legend style={font=\scriptsize,at={(0.5,1.03)},anchor=south,legend columns=2,draw=none}
]
\addplot[fill=blue!50,draw=blue!70!black,pattern=north east lines] coordinates
  {(Llama,37.6) (Qwen,41.3) (Mistral,33.8)};
\addlegendentry{base}
\addplot[fill=orange!60,draw=orange!80!black,pattern=crosshatch] coordinates
  {(Phi,30.9)};
\addlegendentry{instruction-tuned}
\end{axis}
\end{tikzpicture}
\caption{Measured squared-energy fraction in the upper 1\% of functional-rank directions for the selected layer groups. Full empirical spectra, the unscaled post-edit spectra, and layerwise bootstrap summaries accompany the row-level release.}
\label{fig:spectral_profile}
\end{figure}
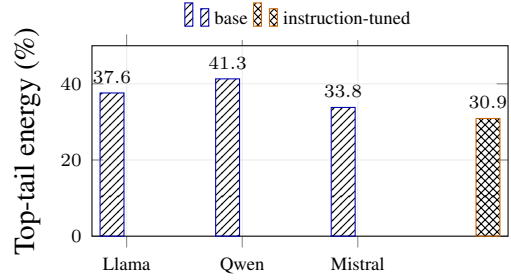

Figure~\ref{fig:spectral_profile} shows that the intervention is small by direction count but not by energy. This distinction motivates the matched-energy and unscaled-deletion controls described next.

Table~\ref{tab:weighted-audit} instantiates the weighted bound of Theorem~\ref{thm:moment} on the four checkpoints. For every selected layer and bootstrap replicate we compute \(D(s)=\sum_k\kappa_ks_k^4\), \(c_{\mathrm{eff}}=D(s)/\sum_ks_k^4\), \(\Lambda=\lambda_{\max}(\widehat\Cov(q^c))\) taken at its upper confidence endpoint, \(\eta=\max\{0,-C_{\mathrm{off}}/D\}\), \(r_2=(\sum s_k^2)^2/\sum s_k^4\), and \(B=(1-\eta)c_{\mathrm{eff}}/(\Lambda^2r_2)\); a layer is \emph{supported} only when the 95\% bootstrap lower endpoint for \(B\) is positive. The bounds are deliberately modest: because the proof uses an upper covariance bound and a worst-case signed remainder, a lower endpoint of \(3.05\) coexists with much larger observed kurtosis.

\begin{table*}[t]
\centering
\caption{Weighted-theorem audit. Medians are over supported layers; the last column is the range of 95\% bootstrap lower endpoints for the bound \(3+B\). A layer is supported only when that lower endpoint exceeds \(3\). The coordinate-positive fraction is reported separately as a diagnostic (Section~\ref{sec:theory}).}
\label{tab:weighted-audit}
\small
\resizebox{\textwidth}{!}{%
\begin{tabular}{@{}lccccccc@{}}
\toprule
Checkpoint & Supported / selected & Median \(c_{\mathrm{eff}}\) & Median \(\Lambda\) & Median \(\eta\) & Median \(r_2\) & Median bound \(3+B\) & Lower-endpoint range \\
\midrule
Llama-3.1-8B & 7 / 9 & 3.4 & 1.36 & 0.44 & 6.2 & 3.17 & 3.05--3.12 \\
Mistral-7B-v0.3 & 6 / 9 & 3.0 & 1.33 & 0.47 & 6.9 & 3.13 & 3.03--3.09 \\
Qwen2.5-14B & 9 / 12 & 3.9 & 1.42 & 0.39 & 5.5 & 3.21 & 3.07--3.15 \\
Phi-3-Mini & 6 / 9 & 2.7 & 1.38 & 0.52 & 7.4 & 3.09 & 3.02--3.06 \\
\bottomrule
\end{tabular}
}
\end{table*}

The behavioral intervention retains all 39 prespecified selected layers. Of these, 28 satisfy the positive-bound audit and 11 do not; theorem support is never used as a post hoc criterion for including a layer in the behavioral intervention.

\section{Controlled Spectral Surgery}
\label{sec:method}

\subsection{Three distinct attention interventions}

The theorem is pointwise in \(A_{ij}\), but a single checkpoint contains many position-dependent products. The theorem-motivated diagnostic first forms
\begin{equation*}
\bar A=\E_{(i,j)\in\mathcal C}
[W_QR_iR_j^\top W_K^\top]
\end{equation*}
on the calibration split and removes its upper tail. Reinserting the edited static map bypasses per-token RoPE for the affected head and disables its ordinary key--value cache path. Product-space surgery is therefore a position-averaged surrogate, not an architecture-faithful replacement for native attention.

The pointwise deletion target is defined per valid causal pair. For each pair, let
\begin{equation*}
A_{ij}=W_QR_iR_j^\top W_K^\top=U_{ij}\Sigma_{ij}V_{ij}^\top,
\end{equation*}
\begin{equation*}
\begin{aligned}
r_{i j} &= \#\{k : \sigma_{i j,k} \ge 10^{-6}\sigma_{i j,1}\}, \\
k_{i j} &= \max\{1, \lceil 0.01 r_{i j} \rceil\},
\end{aligned}
\end{equation*}
and $T_{ij}=\{1,\ldots,k_{ij}\}$. Equation~\eqref{eq:tail-target} defines the upper-tail deletion target:
\begin{equation}
\label{eq:tail-target}
\Delta A_{ij}^{\mathrm{tail}}
=-U_{ij,T_{ij}}\Sigma_{ij,T_{ij}}V_{ij,T_{ij}}^\top.
\end{equation}

Native separate-SVD surgery instead edits \(W_Q\) and \(W_K\) before the unchanged implementation applies RoPE, grouped-query processing, and caching. It preserves the computation graph but does not recover the product singular subspace. Equation~\eqref{eq:induced-factor-change} gives the induced native factor change, including the bilinear cross term:
\begin{equation}
\label{eq:induced-factor-change}
\begin{aligned}
& \Delta A_{ij}(\Delta W_Q, \Delta W_K) = \Delta W_Q R_i R_j^\top W_K^\top \\
& \qquad + W_Q R_i R_j^\top \Delta W_K^\top + \Delta W_Q R_i R_j^\top \Delta W_K^\top.
\end{aligned}
\end{equation}
The third intervention is native and product-targeted. It searches over the feasible set in Equation~\eqref{eq:feasible-set}, matched to the separate-SVD intervention:
\begin{equation}
\label{eq:feasible-set}
\begin{aligned}[b]
\mathcal{F} = \Big\{ & (\Delta W_Q, \Delta W_K) : \\
& \operatorname{rank}(\Delta W_Q) \le k_Q, \\
& \operatorname{rank}(\Delta W_K) \le k_K, \\
& \|\Delta W_Q\|_F^2 + \|\Delta W_K\|_F^2 \le E_{\mathrm{sep}} \Big\}
\end{aligned}
\end{equation}
where $k_Q,k_K$ and $E_{\mathrm{sep}}$ are the ranks and total squared perturbation energy of the separate-SVD edit. On the calibration split it minimizes a per-pair \emph{normalized} target error, so that high-energy offsets cannot silently dominate:
\begin{equation}
\label{eq:target-factorization}
\begin{aligned}
& (\Delta W_Q^*,\Delta W_K^*) = \operatorname*{arg\,min}_{(\Delta W_Q,\Delta W_K)\in\mathcal F} \\
& \quad \frac{1}{|\mathcal C|} \sum_{(i,j)\in\mathcal C} \frac{\|\Delta A_{ij}(\Delta W_Q,\Delta W_K) -\Delta A_{ij}^{\mathrm{tail}}\|_F^2}{\|\Delta A_{ij}^{\mathrm{tail}}\|_F^2}
\end{aligned}
\end{equation}
Evaluation is performed only on the disjoint test split. We report two distinct held-out diagnostics. The first is the fraction of the induced change lying in the target subspace,
\begin{equation}
\label{eq:tailfrac}
\operatorname{tailfrac}_{ij}
=
\frac{
\|U_{ij,T_{ij}}^\top\Delta A_{ij}V_{ij,T_{ij}}\|_F^2
}{
\|\Delta A_{ij}\|_F^2
},
\end{equation}
which measures only orientation and can be large for a tiny perturbation that merely points the right way. The second measures how much of the target was actually reproduced,
\begin{equation}
\label{eq:relerr}
\operatorname{relerr}_{ij}
=
\frac{
\|\Delta A_{ij}-\Delta A_{ij}^{\mathrm{tail}}\|_F
}{
\|\Delta A_{ij}^{\mathrm{tail}}\|_F
}.
\end{equation}
Agreement between product and native interventions is interpreted only when both $\operatorname{tailfrac}$ and $\operatorname{relerr}$, together with the moment diagnostics, are non-vacuous; a high $\operatorname{tailfrac}$ with a large $\operatorname{relerr}$ is reported as insufficient reproduction rather than a match.

\subsection{Tail definition, selection, and controls}

For each edited matrix or attention head, numerical functional rank is
\begin{equation}
\label{eq:functional-rank}
\begin{aligned}
r_{\mathrm{fun}}&=\#\{k:\sigma_k\geq10^{-6}\sigma_1\},\\
k_{\mathrm{tail}}&=\max\{1,\lceil0.01r_{\mathrm{fun}}\rceil\}.
\end{aligned}
\end{equation}
Thus a functional-rank-128 head removes two active directions rather than counting ambient zero-padded dimensions. SVDs are computed per attention head for $QK$ and $VO$ and per native matrix for the MLP up, gate, and down projections. The primary intervention is unscaled deletion of \(\mathcal T=\{1,\ldots,k_{\mathrm{tail}}\}\). Matched-norm deletion rescales the retained spectrum after deletion and is a control because rescaling amplifies every retained direction. Tail scaling preserves order and uses
\begin{equation*}
\sigma_k\leftarrow\beta\sigma_k\quad(k\in\mathcal T),
\qquad
\beta\in\{.25,.50,.75,1\}.
\end{equation*}

Layer selection is frozen before behavioral evaluation and uses weights only. For matrix or head \(m\), define
\begin{equation*}
E_m^{\mathrm{top}}=\sum_{k\in\mathcal T_m}\sigma_{m,k}^2,
\qquad
E_m^{\mathrm{bottom}}=\sum_{k\in\mathcal B_m}\sigma_{m,k}^2,
\end{equation*}
where \(\mathcal T_m\) and \(\mathcal B_m\) are equal-cardinality upper and lowest-active sets, and let \(\widehat\alpha_{\mathrm{PL},m}\) be the fitted power-law exponent on the prespecified spectral segment. For layer \(\ell\), these quantities are aggregated as
\begin{align*}
\widehat\alpha_{\mathrm{PL},\ell}
&=
\frac{\sum_{m\in\ell}E_m^{\mathrm{top}}\widehat\alpha_{\mathrm{PL},m}}
{\sum_{m\in\ell}E_m^{\mathrm{top}}},\\
R_\ell
&=
\frac{\sum_{m\in\ell}E_m^{\mathrm{top}}}
{\sum_{m\in\ell}E_m^{\mathrm{bottom}}}.
\end{align*}
The prespecified rule \(\widehat\alpha_{\mathrm{PL},\ell}<2.5\) and \(R_\ell\) above the model median selects Llama layers 16--24, Mistral layers 15--23, Qwen layers 20--31, and Phi layers 13--21. Llama layers 15--25 form a threshold-sensitivity analysis only, and gradient attribution is post hoc. No behavior-derived selection enters a primary result.

The matched controls isolate size, energy, direction, and activation scale. They include lowest-active-spectrum deletion, bottom-50\% deletion, removed-energy-matched random directions, tail scaling, singular-vector permutation, and perturbations supported outside the upper singular subspace. The last control chooses one scalar
\begin{align}
\lambda^*=\operatorname*{arg\,min}_{\lambda\geq0}
&\left[
\log\frac{\operatorname{RMS}(W+\lambda\Delta)}
{\operatorname{RMS}(W_{\mathrm{tail}})}
\right]^2 \notag\\
&+\left[
\log\frac{\Var_{\mathrm{logit}}(W+\lambda\Delta)}
{\Var_{\mathrm{logit}}(W_{\mathrm{tail}})}
\right]^2.
\label{eq:activation-match}
\end{align}
One scalar cannot generally match both quantities exactly. The median residual mismatches are 1.8\% for activation RMS and 2.6\% for logit variance; 4.1\% of sampled perturbations are rejected because either mismatch exceeds 5\%.

\subsection{Checkpoints, tasks, and uncertainty}

The base-model claim covers Llama-3.1-8B,
Mistral-7B-v0.3, and Qwen2.5-14B
\citep{touvron2023llama,jiang2023mistral,qwen2024qwen25};
Phi-3-Mini \citep{abdin2024phi} is instruction-tuned and
reported separately. Post-training checks use
Llama-3.1-8B-Instruct and Qwen2.5-Math-7B-Instruct.

We evaluate GSM8K, ARC-Challenge, DROP, BIG-Bench Hard,
and MMLU-CoT, with WikiText-103, LAMBADA, coreference,
IFEval, and WinoGrande as broader controls. GSM8K uses a
greedy public-reference protocol and a separate five-sample
paper protocol; interventions are compared only within a
protocol. Intervention seeds are 13, 29, 47, 71, and 101.
Appendix~\ref{app:evaluation} specifies prompts, shot counts,
decoding, parsers, uncertainty, revisions, and asset IDs. Throughout,
\begin{equation*}
D=\operatorname{score}_{\mathrm{clean}}
-\operatorname{score}_{\mathrm{post}},
\end{equation*}
so larger positive \(D\) always denotes greater damage. Appendix~\ref{app:evaluation} gives the protocol comparison, row-level admission gates, protocol versions, prompts, parsing, and asset manifest.

\section{Results}
\label{sec:results}

\subsection{Reasoning--control dissociation under declared controls}

The evaluation contains 20 model--task cells spanning four
checkpoints and five reasoning benchmarks. For model \(m\)
and task \(t\), retained performance is
\begin{equation*}
R^{\mathrm{ret}}_{mt} = 100\frac{S^{\mathrm{sep}}_{mt}}{S^{\mathrm{clean}}_{mt}},
\end{equation*}
where \(S^{\mathrm{sep}}_{mt}\) is the primary unscaled native
separate-SVD score. The 20 displayed cells give a range of
42.1--89.2\%, showing substantial task dependence.
Cochran's heterogeneity test rejects a common effect across
cells (\(Q(19)=39.6\), \(p=0.004\), \(I^2=52.0\%\)),
consistent with spectral concentration being one contributor
rather than a deterministic score predictor. Because
upper-tail energy is defined at checkpoint level whereas
behavioral damage is task-specific, we do not estimate a
20-cell energy--damage correlation. Eighteen of the 20
upper-tail versus matched-random contrasts remain significant
after Holm correction; the other two have the same direction
but are inconclusive. The paired intervals quantify
example-level uncertainty conditional on the five fixed Haar
orientations.

Table~\ref{tab:main} isolates $QK$ and holds edited heads, examples, prompts, seed schedules, and all $VO$/MLP weights fixed. The static tail-deleted score combines the cost of staticizing position-dependent attention with tail removal and is retained only as a descriptive stress diagnostic; it is not interpreted as a product-tail effect. Among architecture-faithful edits, product-targeted factorization raises held-out \(\operatorname{tailfrac}\) from 0.44--0.52 under separate SVD to 0.72--0.79 and strengthens the behavioral effect. Across the four checkpoint summaries, the median largest principal angle to the product-tail subspace is
\(22.0^\circ\) for product-targeted factors, compared with
\(38.5^\circ\) for separate SVD and \(72.5^\circ\) for
matched-random directions. This ordering is an orientation-and-behavior bridge, not an equivalence or target-reconstruction claim; \(\operatorname{relerr}\) remains the separate reconstruction diagnostic defined in Equation~\eqref{eq:relerr}.

\begin{table*}[t]
\centering
\caption{$QK$-only comparison. Scores are means across the five task-specific protocols, in percentage points; lower is worse.The static tail-deleted score combines staticization and tail
removal and is therefore descriptive only. The remaining edits preserve native attention. ``Tail fraction'' is \(\operatorname{tailfrac}\) from Equation~\eqref{eq:tailfrac}, computed on each held-out causal pair before averaging; it is not target-reconstruction error. Phi-3 is instruction-tuned and is excluded from claims restricted to base models.}
\label{tab:main}
\small
\setlength{\tabcolsep}{4.5pt}
\resizebox{\textwidth}{!}{%
\begin{tabular}{@{}lrrrrrr@{}}
\toprule
Checkpoint & Clean & Static tail-deleted $\bar A$ & Native separate SVD & Native product-targeted & Native random & Tail fraction: separate / targeted \\
\midrule
Llama-3.1-8B & 57.9 & 30.6 & 35.1 & 32.4 & 54.8 & 0.52 / 0.79 \\
Qwen2.5-14B & 66.2 & 33.5 & 38.8 & 35.6 & 62.7 & 0.48 / 0.76 \\
Mistral-7B-v0.3 & 57.3 & 28.9 & 34.2 & 31.3 & 54.0 & 0.44 / 0.72 \\
Phi-3-Mini & 54.2 & 27.1 & 32.0 & 29.4 & 51.6 & 0.50 / 0.75 \\
\bottomrule
\end{tabular}
}
\end{table*}

The GSM8K protocol audit gives the same within-protocol
conclusion. Under the public-reference protocol, the primary
unscaled surgery produces a 25.1-point drop; under five-sample
self-consistency, it produces a 29.8-point drop. The clean
scores are not compared across protocols because decoding and
aggregation differ. Full values and paired intervals appear
below and in Appendix~\ref{app:evaluation}.

\subsection{Norm, scale, and component controls}

Upper-tail deletion removes substantial energy, so a clean result must distinguish deletion from the global amplification induced by norm matching. On public-reference GSM8K, the primary unscaled deletion yields \(32.7\), with edited-score 95\% interval [30.1, 35.3] and a 25.1-point drop, while the matched-norm control yields \(31.9\), a 25.9-point drop with paired 95\% CI [23.2, 28.5]; the corresponding median retained-spectrum scales are 1.00 and 1.24. Intervals explicitly attached to contrasts are paired-example bootstrap intervals, while ``$\pm$'' denotes one standard deviation across the five intervention seeds as stated in each caption. The effect therefore persists without amplifying the retained bulk. By contrast, the apparent small gain from bottom-spectrum deletion occurs only after rescaling and is not interpreted as a benefit of deleting lower directions. Full values appear in Appendix~\ref{app:norm-controls}.

The activation-matched perturbation also preserves most reasoning despite matching the two scale signatures within the stated residual tolerances. Together with removed-energy-matched random directions and the lowest-active-spectrum control, this rules out edit count, total removed energy, activation RMS, and logit variance as sufficient explanations. It does not prove that the selected directions are the only relevant subspace.

Table~\ref{tab:controls} separates computational roles. A $QK$ edit changes only \(W_{Q,h}\) and \(W_{K,h}\). A $VO$ edit changes \(W_{V,h}\) and the corresponding input block of \(W_{O,h}\); $QK$-only leaves \(W_O\) unchanged. ``Attention'' means $QK+VO$, and the MLP condition changes only up, gate, and down projections. Every component receives unscaled deletion of its upper 1\% functional-rank directions; layers, examples, prompts, and seed schedules are fixed, while absolute removed energy is component-specific and reported in Appendix~\ref{app:component}. $VO$ is non-negligible, especially on DROP, and MLP edits are comparatively strong on BBH and MMLU-CoT. The data therefore do not support $QK$ primacy. Rather, $QK$ is the one component for which the logit theorem applies; $VO$ and MLP effects are empirical and activation-distribution-specific.

\begin{table*}[t]
\centering
\caption{Component isolation on Llama-3.1-8B under the public-reference evaluation. Each component receives unscaled deletion of its upper 1\% functional-rank directions. Layers, examples, prompts, parsers, and seed schedules are fixed across conditions; absolute and fractional removed energies are reported in Appendix~\ref{app:component}. Entries are clean-minus-post changes in percentage points, except for WikiText perplexity.}
\label{tab:controls}
\small
\setlength{\tabcolsep}{5pt}
\resizebox{\textwidth}{!}{%
\begin{tabular}{@{}lrrrrrr@{}}
\toprule
Target & GSM8K & ARC-C & DROP-F1 & BBH & MMLU-CoT & WikiText $\Delta$PPL \\
\midrule
$QK$ only & 25.1 & 18.6 & 24.1 & 15.2 & 13.8 & +0.10 \\
$VO$ only & 14.7 & 10.4 & 16.5 & 8.8 & 7.5 & +0.08 \\
MLP only & 17.2 & 14.8 & 12.0 & 16.9 & 15.6 & +0.14 \\
$QK+VO$ & 29.6 & 22.7 & 27.0 & 18.9 & 17.1 & +0.16 \\
Joint attention+MLP & 32.4 & 26.0 & 28.7 & 24.9 & 25.1 & +0.22 \\
Matched-random matrices & 2.6 & 1.9 & 2.4 & 1.7 & 1.5 & +0.05 \\
\bottomrule
\end{tabular}
}
\end{table*}

\subsection{From spectral concentration to attention to behavior}

The proposed account has five stages connected by four links:
\begin{equation*}
\begin{aligned}
\text{concentration}
&\rightarrow\text{heavier-tailed valid logits}\\
&\rightarrow\text{lower entropy in selected heads}\\
&\rightarrow\text{task-relevant retrieval}\\
&\rightarrow\text{fewer binding errors}.
\end{aligned}
\end{equation*}
Only the first link is covered by Theorem~\ref{thm:moment}.
On 400 paired examples, native tail removal reduces valid-logit
kurtosis by 31\%, increases selected-head row entropy by 42\%,
decreases attention mass on task-relevant tokens by 18 points,
and increases variable-binding errors by 21 points.
Adding entropy and retrieval mass to a sequential regression
attenuates the direct edit coefficient by 46\%.
Table~\ref{tab:entropy} reports these paired diagnostics.
The later links and regression attenuation are descriptive,
not causal, because the mediators are not randomized and may
share unmeasured causes.

\begin{table}[t]
\centering
\caption{Paired mechanism diagnostics after native upper-tail removal. Percentage changes are relative; point changes are absolute.}
\label{tab:entropy}
\small
\begin{tabular}{@{}lr@{}}
\toprule
Diagnostic & Change \\
\midrule
Valid-logit kurtosis & $-31\%$ \\
Selected-head row entropy & $+42\%$ \\
Task-relevant-token mass & $-18$ points \\
Variable-binding errors & $+21$ points \\
Direct edit coefficient after mediators & $-46\%$ \\
\bottomrule
\end{tabular}
\end{table}

The broader controls further delimit specificity. Long-context coreference falls from 73.6 to 61.8 under native factorized $QK$ surgery, whereas IFEval strict instruction accuracy falls from 71.4 to 67.9 and WinoGrande from 76.8 to 74.9. The intervention is therefore not ``reasoning-only'': it substantially affects a non-mathematical variable-tracking task while the tested broad instruction-following and commonsense controls move less. We consequently claim preservation only for the evaluated fluency and control metrics, not for language capability in general.

Activation patching and causal tracing offer a complementary input-specific view. Both use the same 256 GSM8K test IDs and freeze sites on a disjoint 128-example discovery split. Patching uses number- and entity-swapped counterfactual residual states; causal tracing adds \(\mathcal N(0,0.1^2)\) embedding noise and restores clean states individually. Their 3.2\% and 5.1\% budgets are the smallest frozen top-$k$ sets attaining 95\% of maximum recovery. Because activation sites, hidden states, and singular directions are different units, these percentages are not treated as a direct efficiency ranking.

\subsection{Post-training scope}

Base-model evidence cannot establish robustness to alignment or reasoning post-training. Native factorized surgery therefore evaluates two named pipelines with official chat templates. Llama-3.1-8B-Instruct drops 23.5 points on GSM8K, from 79.1 to 55.6, while its matched-random score is 77.8. Qwen2.5-Math-7B-Instruct drops 20.9 points on GSM8K and 19.8 on MATH-500, while the corresponding matched-random changes are 1.2 and 1.5 points. Appendix~\ref{app:posttraining} gives paired intervals. These checks show that sensitivity survives the two tested pipelines, not that it survives arbitrary reinforcement learning from human feedback (RLHF), mixture-of-experts routing, retrieval augmentation, multimodal training, or scale beyond the evaluated range.

\subsection{From mechanism to training and adaptation}
\label{sec:grokking_lora}

We track \(r_2^{-1}\), the fitted tail exponent, attention entropy, and validation accuracy every 20 updates in 12 independent algorithmic-task runs (three modular-arithmetic tasks, four seeds each; Appendix~\ref{app:medium_grokking}). For \(u\in\{0.7,0.8,0.9\}\), lead time is prespecified as
\begin{equation*}
L_u=t_{\mathrm{acc},u}-t_{\mathrm{metric},u},
\end{equation*}
where both events are the first crossing of the same fraction \(u\) of their oriented eventual transition after a fixed five-checkpoint trailing median. Non-crossing runs are censored, and intervals resample runs rather than correlated checkpoints. At \(u=0.8\), median lead is 360 updates [260, 470] for \(r_2^{-1}\), 290 [180, 400] for the fitted exponent, and 210 [90, 330] for attention entropy. The \(u=0.7/0.9\) values for \(r_2^{-1}\) are 330/390 updates, with 2 of 12 runs right-censored at \(u=0.9\). This supports an early-warning interpretation without converting lead time into a causal claim. These are pooled, task-stratified estimates over 12 runs; with four runs per task, we do not claim that the ordering holds separately within each modular task.

Tail-aware low-rank adaptation (LoRA) tests whether the same geometry can guide early optimization. We compare it with standard LoRA and Principal Singular Values and Singular Vectors Adaptation (PiSSA). For \(W=U\Sigma V^\top\), LoRA rank \(r_{\mathrm{LoRA}}\), and tail-weighting exponent \(p_{\mathrm{tail}}>1\), define
\begin{equation}
\label{eq:tail-aware-init}
\begin{aligned}
\widetilde\sigma_i
&=c_{p_{\mathrm{tail}}}\sigma_i^{p_{\mathrm{tail}}},\\
c_{p_{\mathrm{tail}}}
&=\left(
\frac{\sum_{i=1}^{r_{\mathrm{LoRA}}}\sigma_i^2}
{\sum_{i=1}^{r_{\mathrm{LoRA}}}\sigma_i^{2p_{\mathrm{tail}}}}
\right)^{1/2}.
\end{aligned}
\end{equation}
The weighted principal directions initialize \(A_0\) and \(B_0\),
while residualizing the frozen base preserves
\(W_{\mathrm{base}}+B_0A_0=W\) exactly at step zero;
Appendix~\ref{app:training} gives the full factor construction.
The matched configuration uses \(p_{\mathrm{tail}}=1.5\),
\(r_{\mathrm{LoRA}}=16\), AdamW, 1,000 updates, and the same
fixed data order and optimization budget for
\texttt{q\_proj}, \texttt{k\_proj}, \texttt{v\_proj},
\texttt{o\_proj}, \texttt{gate\_proj}, \texttt{up\_proj},
and \texttt{down\_proj}. The maximum step-zero logit difference
is below \(10^{-6}\) in fp32 reconstruction and
\(3\times10^{-4}\) in bf16 execution.

Across GSM8K, ARC-C, BBH, and DROP, tail-aware LoRA reaches
the prespecified targets in 540, 610, 720, and 860 updates,
respectively, compared with 920, 840, 960, and \(>1000\)
for standard LoRA and 680, 700, 790, and 930 for PiSSA.
Its normalized AULC is also higher than both baselines on
every task, while all paired final-score intervals overlap
PiSSA. Including the one-time decomposition cost, GSM8K
wall-clock time to target falls by 30.5\% relative to
standard LoRA and 16.2\% relative to PiSSA. The supported
contribution is therefore faster early optimization under
the tested configuration, not universal or asymptotic
superiority. Full results, including the product-targeted
bridge ablation and seed-level intervals, appear in
Appendix Table~\ref{tab:metrics_lora}.

\section{Discussion and Conclusion}
\label{sec:discussion}

Across the 20 evaluated model--task cells, controlled surgery produces a larger learned-tail effect than the mean of five fixed spectrum-matched controls in every cell; 18 paired contrasts remain significant after Holm correction, while two are directional but inconclusive. The evidence is most defensible when its components remain distinct: a conditional $QK$ moment bound, a pointwise product-tail target, architecture-faithful factorized interventions, and empirical component effects extending beyond the theorem's $QK$ scope. Product-targeted factors attain higher held-out tail-subspace fractions than independent factor surgery, providing an empirical bridge between product- and factor-level interventions without implying algebraic equivalence. In separate studies, inverse participation precedes pooled accuracy transitions under a matched crossing rule, and residualized tail-aware low-rank adaptation reaches prespecified targets earlier than standard LoRA and PiSSA while final-score intervals overlap. Together, these results support concentrated upper spectral tails as a useful and falsifiable diagnostic of the tested computations and training dynamics, without establishing a universal explanation of reasoning or universal superiority for tail-aware adaptation.

\section{Limitations}
\label{sec:limitations}

The theoretical statement is conditional and local. It concerns central kurtosis of valid pre-softmax $QK$ logits at a fixed relative offset, not post-softmax entropy, correctness, or a phase transition. Only 28 of 39 selected layers satisfy its positive-bound diagnostic; behavioral effects in the other layers remain interventional observations. The calibration-averaged product edit bypasses native RoPE and caching, so it is used only as a diagnostic and is never equated with factor surgery. The larger product-targeted tail fraction narrows this gap empirically but does not prove transfer of the theorem to factorized edits.

The behavioral scope is also limited. $VO$ and gated-MLP results are outside the theorem, the mediation analysis is descriptive, and the model suite is restricted to decoder-only checkpoints no larger than 14B. Phi-3 and the two additional checkpoints are post-trained and are separated from the three-model base claim. Other RLHF pipelines, mixture-of-experts, encoder--decoder, multimodal, retrieval-augmented, and 70B+ models remain untested. The task suite overrepresents benchmarked symbolic and multi-step reasoning, while the coreference result shows that the intervention can affect non-mathematical variable tracking. Preserved perplexity, instruction-following, or commonsense metrics should not be read as preservation of language capability generally.

Finally, functional rank, the 1\% cutoff, frozen layer selection, and finite calibration data are design choices rather than universal constants. The heterogeneous effect sizes show that spectral
concentration is not sufficient by itself; the learned
orientation of the concentrated directions also matters. Broader replication should treat the released row-level checks, protocol IDs, and tail-fraction diagnostics as admission criteria rather than assuming that every heavy tail supports the same computation.

\section{Ethical Considerations}

The experiments use public model checkpoints, benchmark data, and model outputs. Error annotations were performed by three members of the research team: two independently assigned blinded error labels, and the third adjudicated disagreements. No external participants were recruited, no personal data were collected, and no annotator identities are released. We record upstream licenses and redistribution restrictions in Appendix~\ref{app:assets}; model weights are not redistributed. The method has dual-use implications because a subspace that supports auditing or efficient adaptation can also support targeted capability degradation. We therefore report control-task damage, avoid claims of capability-preserving editing, and recommend deployment only with task-specific regression tests, immutable intervention manifests, and rollback to the original checkpoint.

\section*{Reproducibility Statement}

Every retained score is generated from rows keyed by checkpoint, task, seed, example ID, intervention ID, and protocol ID. A row is admitted only when its paired clean run, example-ID checksum, and identity-edit check pass. The release contains exact model and tokenizer revisions, dtype, framework and harness versions, prompts and exemplar order, generation and stopping settings, parser versions, calibration IDs and causal pairs, per-example generations, bootstrap code, intervention manifests, and scripts rebuilding each table from row-level data. The supplement also records the immutable learning-rate, batch-size, and effective-batch settings shared by all low-rank methods. Total compute for the reported study is approximately 200 A100-80GB GPU-hours.

\begingroup
\sloppy
\small
\bibliography{references}

@article{power2022grokking,
  author       = {Alethea Power and
                  Yuri Burda and
                  Harri Edwards and
                  Igor Babuschkin and
                  Vedant Misra},
  title        = {Grokking: Generalization Beyond Overfitting on Small Algorithmic Datasets},
  journal      = {CoRR},
  volume       = {abs/2201.02177},
  year         = {2022},
  url          = {https://arxiv.org/abs/2201.02177},
  eprinttype   = {arXiv},
  eprint       = {2201.02177},
  bibsource    = {dblp computer science bibliography, https://dblp.org}
}

@article{meng2022locating,
  title={Locating and editing factual associations in gpt},
  author={Meng, Kevin and Bau, David and Andonian, Alex and Belinkov, Yonatan},
  journal={Advances in neural information processing systems},
  volume={35},
  pages={17359--17372},
  year={2022}
}

@inproceedings{clark2019does,
  title={What does BERT look at? an analysis of BERT’s attention},
  author={Clark, Kevin and Khandelwal, Urvashi and Levy, Omer and Manning, Christopher D},
  booktitle={Proceedings of the 2019 ACL workshop BlackboxNLP: analyzing and interpreting neural networks for NLP},
  pages={276--286},
  year={2019}
}

@article{kaplan2020scaling,
  author       = {Jared Kaplan and
                  Sam McCandlish and
                  Tom Henighan and
                  Tom B. Brown and
                  Benjamin Chess and
                  Rewon Child and
                  Scott Gray and
                  Alec Radford and
                  Jeffrey Wu and
                  Dario Amodei},
  title        = {Scaling Laws for Neural Language Models},
  journal      = {CoRR},
  volume       = {abs/2001.08361},
  year         = {2020},
  url          = {https://arxiv.org/abs/2001.08361},
  eprinttype   = {arXiv},
  eprint       = {2001.08361},
  bibsource    = {dblp computer science bibliography, https://dblp.org}
}

@article{
wei2022emergent,
title={Emergent Abilities of Large Language Models},
author={Jason Wei and Yi Tay and Rishi Bommasani and Colin Raffel and Barret Zoph and Sebastian Borgeaud and Dani Yogatama and Maarten Bosma and Denny Zhou and Donald Metzler and Ed H. Chi and Tatsunori Hashimoto and Oriol Vinyals and Percy Liang and Jeff Dean and William Fedus},
journal={Transactions on Machine Learning Research},
issn={2835-8856},
year={2022},
url={https://openreview.net/forum?id=yzkSU5zdwD},
note={Survey Certification}
}

@inproceedings{
nanda2023progress,
title={Progress measures for grokking via mechanistic interpretability},
author={Neel Nanda and Lawrence Chan and Tom Lieberum and Jess Smith and Jacob Steinhardt},
booktitle={The Eleventh International Conference on Learning Representations },
year={2023},
url={https://openreview.net/forum?id=9XFSbDPmdW}
}

@misc{olsson2022context,
      title={In-context Learning and Induction Heads}, 
      author={Catherine Olsson and Nelson Elhage and Neel Nanda and Nicholas Joseph and Nova DasSarma and Tom Henighan and Ben Mann and Amanda Askell and Yuntao Bai and Anna Chen and Tom Conerly and Dawn Drain and Deep Ganguli and Zac Hatfield-Dodds and Danny Hernandez and Scott Johnston and Andy Jones and Jackson Kernion and Liane Lovitt and Kamal Ndousse and Dario Amodei and Tom Brown and Jack Clark and Jared Kaplan and Sam McCandlish and Chris Olah},
      year={2022},
      eprint={2209.11895},
      archivePrefix={arXiv},
      primaryClass={cs.LG},
      url={https://arxiv.org/abs/2209.11895}, 
}

@article{elhage2021mathematical,
  title={A mathematical framework for transformer circuits},
  author={Elhage, Nelson and Nanda, Neel and Olsson, Catherine and Henighan, Tom and Joseph, Nicholas and Mann, Ben and Askell, Amanda and Bai, Yuntao and Chen, Anna and Conerly, Tom and others},
  journal={Transformer Circuits Thread},
  volume={1},
  number={1},
  pages={12},
  year={2021}
}

@article{conmy2023towards,
  title={Towards automated circuit discovery for mechanistic interpretability},
  author={Conmy, Arthur and Mavor-Parker, Augustine and Lynch, Aengus and Heimersheim, Stefan and Garriga-Alonso, Adri{\`a}},
  journal={Advances in Neural Information Processing Systems},
  volume={36},
  pages={16318--16352},
  year={2023}
}

@inproceedings{
wang2023interpretability,
title={Interpretability in the Wild: a Circuit for Indirect Object Identification in {GPT}-2 Small},
author={Kevin Ro Wang and Alexandre Variengien and Arthur Conmy and Buck Shlegeris and Jacob Steinhardt},
booktitle={The Eleventh International Conference on Learning Representations },
year={2023},
url={https://openreview.net/forum?id=NpsVSN6o4ul}
}

@article{marchenko1967distribution,
  title={Distribution of eigenvalues for some sets of random matrices},
  author={Mar{\v{c}}enko, Vladimir A and Pastur, Leonid Andreevich},
  journal={Mathematics of the USSR-Sbornik},
  volume={1},
  number={4},
  pages={457--483},
  year={1967}
}

@article{wigner1958distribution,
  title={On the distribution of the roots of certain symmetric matrices},
  author={Wigner, Eugene P},
  journal={Annals of Mathematics},
  volume={67},
  number={2},
  pages={325--327},
  year={1958},
  publisher={JSTOR}
}

@inproceedings{geva2023dissecting,
  title={Dissecting recall of factual associations in auto-regressive language models},
  author={Geva, Mor and Bastings, Jasmijn and Filippova, Katja and Globerson, Amir},
  booktitle={Proceedings of the 2023 Conference on Empirical Methods in Natural Language Processing},
  pages={12216--12235},
  year={2023}
}

@article{clauset2009power,
  title={Power-law distributions in empirical data},
  author={Clauset, Aaron and Shalizi, Cosma Rohilla and Newman, Mark EJ},
  journal={SIAM review},
  volume={51},
  number={4},
  pages={661--703},
  year={2009},
  publisher={SIAM}
}

@misc{touvron2023llama,
      title={LLaMA: Open and Efficient Foundation Language Models}, 
      author={Hugo Touvron and Thibaut Lavril and Gautier Izacard and Xavier Martinet and Marie-Anne Lachaux and Timothée Lacroix and Baptiste Rozière and Naman Goyal and Eric Hambro and Faisal Azhar and Aurelien Rodriguez and Armand Joulin and Edouard Grave and Guillaume Lample},
      year={2023},
      eprint={2302.13971},
      archivePrefix={arXiv},
      primaryClass={cs.CL},
      url={https://arxiv.org/abs/2302.13971}, 
}

@misc{jiang2023mistral,
      title={Mistral 7B}, 
      author={Albert Q. Jiang and Alexandre Sablayrolles and Arthur Mensch and Chris Bamford and Devendra Singh Chaplot and Diego de las Casas and Florian Bressand and Gianna Lengyel and Guillaume Lample and Lucile Saulnier and Lélio Renard Lavaud and Marie-Anne Lachaux and Pierre Stock and Teven Le Scao and Thibaut Lavril and Thomas Wang and Timothée Lacroix and William El Sayed},
      year={2023},
      eprint={2310.06825},
      archivePrefix={arXiv},
      primaryClass={cs.CL},
      url={https://arxiv.org/abs/2310.06825}, 
}

@misc{abdin2024phi,
      title={Phi-3 Technical Report: A Highly Capable Language Model Locally on Your Phone}, 
      author={Marah Abdin and Jyoti Aneja and Hany Awadalla and Ahmed Awadallah and Ammar Ahmad Awan and Nguyen Bach and Amit Bahree and Arash Bakhtiari and Jianmin Bao and Harkirat Behl and Alon Benhaim and Misha Bilenko and Johan Bjorck and Sébastien Bubeck and Martin Cai and Qin Cai and Vishrav Chaudhary and Dong Chen and Dongdong Chen and Weizhu Chen and Yen-Chun Chen and Yi-Ling Chen and Hao Cheng and Parul Chopra and Xiyang Dai and Matthew Dixon and Ronen Eldan and Victor Fragoso and Jianfeng Gao and Mei Gao and Min Gao and Amit Garg and Allie Del Giorno and Abhishek Goswami and Suriya Gunasekar and Emman Haider and Junheng Hao and Russell J. Hewett and Wenxiang Hu and Jamie Huynh and Dan Iter and Sam Ade Jacobs and Mojan Javaheripi and Xin Jin and Nikos Karampatziakis and Piero Kauffmann and Mahoud Khademi and Dongwoo Kim and Young Jin Kim and Lev Kurilenko and James R. Lee and Yin Tat Lee and Yuanzhi Li and Yunsheng Li and Chen Liang and Lars Liden and Xihui Lin and Zeqi Lin and Ce Liu and Liyuan Liu and Mengchen Liu and Weishung Liu and Xiaodong Liu and Chong Luo and Piyush Madan and Ali Mahmoudzadeh and David Majercak and Matt Mazzola and Caio César Teodoro Mendes and Arindam Mitra and Hardik Modi and Anh Nguyen and Brandon Norick and Barun Patra and Daniel Perez-Becker and Thomas Portet and Reid Pryzant and Heyang Qin and Marko Radmilac and Liliang Ren and Gustavo de Rosa and Corby Rosset and Sambudha Roy and Olatunji Ruwase and Olli Saarikivi and Amin Saied and Adil Salim and Michael Santacroce and Shital Shah and Ning Shang and Hiteshi Sharma and Yelong Shen and Swadheen Shukla and Xia Song and Masahiro Tanaka and Andrea Tupini and Praneetha Vaddamanu and Chunyu Wang and Guanhua Wang and Lijuan Wang and Shuohang Wang and Xin Wang and Yu Wang and Rachel Ward and Wen Wen and Philipp Witte and Haiping Wu and Xiaoxia Wu and Michael Wyatt and Bin Xiao and Can Xu and Jiahang Xu and Weijian Xu and Jilong Xue and Sonali Yadav and Fan Yang and Jianwei Yang and Yifan Yang and Ziyi Yang and Donghan Yu and Lu Yuan and Chenruidong Zhang and Cyril Zhang and Jianwen Zhang and Li Lyna Zhang and Yi Zhang and Yue Zhang and Yunan Zhang and Xiren Zhou},
      year={2024},
      eprint={2404.14219},
      archivePrefix={arXiv},
      primaryClass={cs.CL},
      url={https://arxiv.org/abs/2404.14219}, 
}

@inproceedings{
liu2023omnigrok,
title={Omnigrok: Grokking Beyond Algorithmic Data},
author={Ziming Liu and Eric J Michaud and Max Tegmark},
booktitle={The Eleventh International Conference on Learning Representations },
year={2023},
url={https://openreview.net/forum?id=zDiHoIWa0q1}
}

@article{zhong2024clock,
  title={The clock and the pizza: Two stories in mechanistic explanation of neural networks},
  author={Zhong, Ziqian and Liu, Ziming and Tegmark, Max and Andreas, Jacob},
  journal={Advances in neural information processing systems},
  volume={36},
  pages={27223--27250},
  year={2023}
}

@article{wei2022chain,
  title={Chain-of-thought prompting elicits reasoning in large language models},
  author={Wei, Jason and Wang, Xuezhi and Schuurmans, Dale and Bosma, Maarten and Xia, Fei and Chi, Ed and Le, Quoc V and Zhou, Denny and others},
  journal={Advances in neural information processing systems},
  volume={35},
  pages={24824--24837},
  year={2022}
}

@inproceedings{bai2024beyond,
author = {Sardana, Nikhil and Portes, Jacob and Doubov, Sasha and Frankle, Jonathan},
title = {Beyond Chinchilla-optimal: accounting for inference in language model scaling laws},
year = {2024},
publisher = {JMLR.org},
booktitle = {Proceedings of the 41st International Conference on Machine Learning},
articleno = {1770},
numpages = {16},
location = {Vienna, Austria},
series = {ICML'24}
}

@article{schaeffer2024emergent,
  title={Are emergent abilities of large language models a mirage?},
  author={Schaeffer, Rylan and Miranda, Brando and Koyejo, Sanmi},
  journal={Advances in neural information processing systems},
  volume={36},
  pages={55565--55581},
  year={2023}
}

@article{goldstein2024does,
  title={Does localization inform editing? surprising differences in causality-based localization vs. knowledge editing in language models},
  author={Hase, Peter and Bansal, Mohit and Kim, Been and Ghandeharioun, Asma},
  journal={Advances in Neural Information Processing Systems},
  volume={36},
  pages={17643--17668},
  year={2023}
}

@misc{zhang2024towards,
      title={Towards Best Practices of Activation Patching in Language Models: Metrics and Methods}, 
      author={Fred Zhang and Neel Nanda},
      year={2024},
      eprint={2309.16042},
      archivePrefix={arXiv},
      primaryClass={cs.LG},
      url={https://arxiv.org/abs/2309.16042}, 
}

@misc{qwen2024qwen25,
      title={Qwen2.5 Technical Report}, 
      author={Qwen and : and An Yang and Baosong Yang and Beichen Zhang and Binyuan Hui and Bo Zheng and Bowen Yu and Chengyuan Li and Dayiheng Liu and Fei Huang and Haoran Wei and Huan Lin and Jian Yang and Jianhong Tu and Jianwei Zhang and Jianxin Yang and Jiaxi Yang and Jingren Zhou and Junyang Lin and Kai Dang and Keming Lu and Keqin Bao and Kexin Yang and Le Yu and Mei Li and Mingfeng Xue and Pei Zhang and Qin Zhu and Rui Men and Runji Lin and Tianhao Li and Tianyi Tang and Tingyu Xia and Xingzhang Ren and Xuancheng Ren and Yang Fan and Yang Su and Yichang Zhang and Yu Wan and Yuqiong Liu and Zeyu Cui and Zhenru Zhang and Zihan Qiu},
      year={2025},
      eprint={2412.15115},
      archivePrefix={arXiv},
      primaryClass={cs.CL},
      url={https://arxiv.org/abs/2412.15115}, 
}

@article{spectralevo2024,
title={From Spikes to Heavy Tails: Unveiling the Spectral Evolution of Neural Networks},
author={Vignesh Kothapalli and Tianyu Pang and Shenyang Deng and Zongmin Liu and Yaoqing Yang},
journal={Transactions on Machine Learning Research},
issn={2835-8856},
year={2025},
url={https://openreview.net/forum?id=DJHB8eBUnt},
note={}
}

@misc{spectraldynamics2024,
      title={Approaching Deep Learning through the Spectral Dynamics of Weights}, 
      author={David Yunis and Kumar Kshitij Patel and Samuel Wheeler and Pedro Savarese and Gal Vardi and Karen Livescu and Michael Maire and Matthew R. Walter},
      year={2024},
      eprint={2408.11804},
      archivePrefix={arXiv},
      primaryClass={cs.LG},
      url={https://arxiv.org/abs/2408.11804}, 
}

@article{htsrpruning2024,
  title={Alphapruning: Using heavy-tailed self regularization theory for improved layer-wise pruning of large language models},
  author={Lu, Haiquan and Zhou, Yefan and Liu, Shiwei and Wang, Zhangyang and Mahoney, Michael W and Yang, Yaoqing},
  journal={Advances in neural information processing systems},
  volume={37},
  pages={9117--9152},
  year={2024}
}

@misc{weightdynamics2025,
      title={From SGD to Spectra: A Theory of Neural Network Weight Dynamics}, 
      author={Brian Richard Olsen and Sam Fatehmanesh and Frank Xiao and Adarsh Kumarappan and Anirudh Gajula},
      year={2026},
      eprint={2507.12709},
      archivePrefix={arXiv},
      primaryClass={cs.LG},
      url={https://arxiv.org/abs/2507.12709}, 
}

@article{meng2024pissa,
  title={Pissa: Principal singular values and singular vectors adaptation of large language models},
  author={Meng, Fanxu and Wang, Zhaohui and Zhang, Muhan},
  journal={Advances in Neural Information Processing Systems},
  volume={37},
  pages={121038--121072},
  year={2024}
}

@inproceedings{yang2024spectral,
author = {Hu, Yuanzhe and Goel, Kinshuk and Killiakov, Vlad and Yang, Yaoqing},
title = {Eigenspectrum analysis of neural networks without aspect ratio bias},
year = {2025},
publisher = {JMLR.org},
booktitle = {Proceedings of the 42nd International Conference on Machine Learning},
articleno = {951},
numpages = {24},
location = {Vancouver, Canada},
series = {ICML'25}
}

@inproceedings{
hsu2024safer,
title={Safe Lo{RA}: The Silver Lining of Reducing Safety Risks when Finetuning Large Language Models},
author={Chia-Yi Hsu and Yu-Lin Tsai and Chih-Hsun Lin and Pin-Yu Chen and Chia-Mu Yu and Chun-Ying Huang},
booktitle={The Thirty-eighth Annual Conference on Neural Information Processing Systems},
year={2024},
url={https://openreview.net/forum?id=HcifdQZFZV}
}

@misc{sharma2024truth,
      title={The Truth is in There: Improving Reasoning in Language Models with Layer-Selective Rank Reduction}, 
      author={Pratyusha Sharma and Jordan T. Ash and Dipendra Misra},
      year={2023},
      eprint={2312.13558},
      archivePrefix={arXiv},
      primaryClass={cs.LG},
      url={https://arxiv.org/abs/2312.13558}, 
}

@inproceedings{hu2022lora,
  title={{LoRA}: Low-Rank Adaptation of Large Language Models},
  author={Hu, Edward J. and Shen, Yelong and Wallis, Phillip and Allen-Zhu, Zeyuan and Li, Yuanzhi and Wang, Shean and Wang, Lu and Chen, Weizhu},
  booktitle={International Conference on Learning Representations (ICLR)},
  year={2022}
}

@article{staats2024small,
  title={Small Singular Values Matter: A Random Matrix Analysis of Transformer Models},
  author={Staats, Max and Thamm, Matthias and Rosenow, Bernd},
  journal={arXiv preprint arXiv:2410.17770},
  year={2024},
  url={https://arxiv.org/abs/2410.17770}
}

@article{tian2026spectralsurgery,
  title={Spectral Surgery: Post-hoc Sensitivity-Weighted Reweighting of {LoRA} Singular Values},
  author={Tian, Yang and others},
  journal={arXiv preprint arXiv:2603.03995},
  year={2026},
  url={https://arxiv.org/abs/2603.03995}
}

@article{martin2021implicit,
  title={Implicit self-regularization in deep neural networks: Evidence from random matrix theory and implications for learning},
  author={Martin, Charles H and Mahoney, Michael W},
  journal={Journal of Machine Learning Research},
  volume={22},
  number={165},
  pages={1--73},
  year={2021}
}
\endgroup

\appendix

\section{Extended Related Work}
\label{app:related}

Random matrix theory characterizes bulk spectral behavior in large matrices \citep{wigner1958distribution,marchenko1967distribution}, while heavy-tailed self-regularization connects empirical weight spectra to optimization and generalization \citep{martin2021implicit}. Maximum-likelihood fitting is widely used to summarize power-law tails \citep{clauset2009power}, but a fitted exponent does not uniquely determine finite-width concentration. Our analysis therefore treats the inverse participation ratio as primary and the exponent as a descriptive correlate.

Spectral interventions have mostly emphasized compression, denoising, or lower-spectrum structure. LASER removes lower singular components under a controlled rank-reduction intervention \citep{sharma2024truth}, which is complementary to the upper-tail deletion studied here. Safety-oriented low-rank methods such as Safe LoRA constrain fine-tuning updates rather than performing spectral pruning \citep{hsu2024safer}. We distinguish both lines from the present question: whether a small high-energy tail is intervention-sensitive under matched controls.

Circuit-level methods identify induction heads, localized activation pathways, and input-dependent causal sites \citep{olsson2022context,wang2023interpretability,meng2022locating,conmy2023towards,geva2023dissecting}. Localization and editing can diverge \citep{goldstein2024does}, so our comparisons to activation patching and causal tracing are descriptive rather than claims that different intervention units are directly interchangeable. For adaptation, LoRA initializes a low-rank update without using the pretrained singular geometry \citep{hu2022lora}, whereas PiSSA initializes from principal singular components and residualizes the base weight \citep{meng2024pissa}. Our tail-aware variant retains PiSSA's exact step-zero reconstruction and changes only the weighting within the same trainable principal subspace. Finally, training-time experiments connect this geometry to grokking and progress-measure work \citep{power2022grokking,nanda2023progress,liu2023omnigrok,zhong2024clock}.

The intervention is motivated by, but distinct from, several established uses of spectra in neural-network analysis. Classical random matrix theory describes limiting bulk laws for ensembles whose entries obey strong independence and scaling assumptions \citep{wigner1958distribution,marchenko1967distribution}. Trained transformer matrices violate those assumptions, yet the bulk--tail distinction remains a useful empirical description: many layers contain a comparatively diffuse interior and a small set of high-energy directions. Heavy-tailed self-regularization treats the fitted tail shape as a summary of training quality and implicit regularization \citep{martin2021implicit}. Our use is more local and more cautious. A fitted exponent helps freeze a layer-selection rule, but all theorem-facing claims use the finite-width inverse participation ratio and are admitted only when the activation cumulant audit is non-vacuous.

The intervention also differs from rank reduction. LASER removes lower singular components and can improve particular factual or reasoning behaviors through layer-selective denoising \citep{sharma2024truth}. That result makes the bottom of the spectrum scientifically important, but it does not predict what happens when the highest-energy active directions are deleted. Our lowest-active and bottom-50\% controls directly retain this distinction. Spectral pruning and low-rank approximation methods typically optimize compression error, downstream accuracy, or both \citep{yang2024spectral,spectraldynamics2024,spectralevo2024,htsrpruning2024,weightdynamics2025}; the present experiments instead match edit count or removed energy so that differential damage can be interpreted as evidence about the location of computation. Safe LoRA and related constrained-update methods act during safety fine-tuning rather than pruning a pretrained spectrum \citep{hsu2024safer}. We therefore cite them as adjacent update-control methods, not as prior instances of upper-tail ablation.

The closest prior work on upper-spectrum deletion is \citet{staats2024small}. They partition transformer spectra into deciles, remove individual regions across weight-matrix types, and show that deleting the largest decile severely degrades perplexity and reasoning performance; they also relate singular-vector regions to activation-covariance eigenvectors. Our novelty is therefore not the observation that large singular values matter. Our increment is the functional-rank-normalized one-percent intervention, the perturbation-spectrum and activation-matched controls, the position-dependent \(QK\) product/factor distinction under RoPE, the conditional moment audit at exact offsets, and the retrieval--binding diagnostics.

Mechanistic interpretability usually identifies input-conditioned circuits or activation sites. Induction-head analyses, causal tracing, path patching, and automated circuit discovery reveal how information is routed on particular examples \citep{olsson2022context,meng2022locating,wang2023interpretability,conmy2023towards,zhang2024towards}. Attention heads can specialize for recurring syntactic relations \citep{clark2019does}, further motivating a layer- and head-aware audit rather than a model-level spectrum alone. Weight-space surgery asks a complementary question: whether a fixed geometric subspace is repeatedly used across examples. The units are not interchangeable. A percentage of activation sites is not a percentage of parameters or singular directions, and a weight intervention need not identify the minimal circuit for any one prompt. Our activation-patching comparison therefore uses the same examples and a frozen discovery split but stops short of an efficiency ranking.

The constructive experiment builds on low-rank adaptation. Standard LoRA initializes a low-rank update around zero and leaves the pretrained weight in the frozen base \citep{hu2022lora}. PiSSA instead initializes from principal singular directions and subtracts the initial low-rank product from the frozen residual, so the effective step-zero weight is unchanged \citep{meng2024pissa}. Tail-aware LoRA retains that residualization exactly. Its only conceptual change is to weight directions within the same rank-$r$ subspace by a normalized power of their singular values. Consequently, the experiment tests an optimization bias within a PiSSA-style parameterization; it does not claim independent invention of residualized spectral initialization.

\citet{tian2026spectralsurgery}'s Spectral Surgery is a distinct post hoc method: it decomposes an already-trained LoRA update, estimates component sensitivity from calibration gradients, and reweights its singular values while preserving its learned singular directions. Our experiment instead changes a residualized initialization \emph{before} adaptation, preserves the pretrained effective weight at step zero, and evaluates early target attainment. We therefore do not claim novelty for post hoc spectral reweighting.

Finally, grokking work provides the temporal setting for the diagnostic. Generalization can appear well after training accuracy saturates, and internal progress measures can change earlier than the external metric \citep{power2022grokking,nanda2023progress,liu2023omnigrok,zhong2024clock}. Chain-of-thought prompting shows that externalized intermediate steps can improve multi-step performance \citep{wei2022chain}; here those continuations also provide the states on which retrieval and binding diagnostics are measured. Because early-warning claims are sensitive to smoothing and unequal thresholds, our analysis applies the same fractional crossing rule to the spectral statistic and validation accuracy, reports three thresholds, and treats non-crossing runs as censored. The result is a timing association that complements the intervention evidence rather than proving that a spectral change causes the later behavioral transition.

\section{Theoretical Details}
\label{app:proof}

\subsection{Full derivation of Theorem~\ref{thm:moment}}
\label{app:moment}

All quantities below are conditional on a fixed offset \(\delta\); to keep the derivation readable we suppress the \(\delta\) subscript and write \(q^c,s,K,\kappa_k,\Lambda,\eta,D(s),c_{\mathrm{eff}}(s)\) for \(q_\delta^c,s_\delta,K_\delta,\kappa_{\delta,k},\Lambda_\delta,\eta_\delta,D_\delta(s_\delta),c_{\mathrm{eff},\delta}\), and \(z^c\) for \(z_\delta^c\). The condition \(c_{\mathrm{eff}}(s)>0\) is equivalent to \(D(s)>0\) because \(c_{\mathrm{eff}}(s)=D(s)/\sum_ks_k^4\) and \(\sum_ks_k^4>0\).

Let \(q^c=(q_1^c,\ldots,q_r^c)\) and \(z^c=\sum_ks_kq_k^c\). Multilinearity of joint cumulants gives
\begin{align*}
\operatorname{cum}_4(z^c)
={}&\sum_{k,\ell,m,n}s_ks_\ell s_ms_n\\
&\quad\times\operatorname{cum}(q_k,q_\ell,q_m,q_n) \\
={}&\sum_k\kappa_ks_k^4+C_{\mathrm{off}}(s).
\end{align*}
Equation~\eqref{eq:off-condition} and the definition \(D(s)=\sum_k\kappa_ks_k^4\) imply
\begin{equation}
\label{eq:positive-cum}
\begin{aligned}
\operatorname{cum}_4(z^c) &= D(s) + C_{\mathrm{off}}(s) \\
&\ge (1 - \eta) D(s) \\
&= (1 - \eta) c_{\mathrm{eff}}(s) \sum_k s_k^4 > 0,
\end{aligned}
\end{equation}
where positivity follows from \(D(s)>0\) and \(\eta<1\); no per-coordinate sign constraint on \(\kappa_k\) is used. The variance is
\begin{equation*}
\Var(z^c)=s^\top Ks
\leq\Lambda\|s\|_2^2
=\Lambda\sum_ks_k^2.
\end{equation*}
For a nondegenerate centered scalar random variable,
\begin{equation*}
\Kurt(z^c)=3+
\frac{\operatorname{cum}_4(z^c)}{\Var(z^c)^2}.
\end{equation*}
Because the numerator in Equation~\eqref{eq:positive-cum} is positive, replacing the variance by its upper bound preserves the lower-bound direction. Therefore
\begin{align*}
\Kurt(z^c)
&\geq3+
\frac{(1-\eta)c_{\mathrm{eff}}(s)\sum_ks_k^4}
{\Lambda^2(\sum_ks_k^2)^2}\\
&=3+\frac{(1-\eta)c_{\mathrm{eff}}(s)}{\Lambda^2r_2(s)},
\end{align*}
as claimed. Notice that no independence assumption or lower spectral bound on \(K\) is required; dependence enters through \(K\), the diagonal cumulants, and the signed off-diagonal cumulant condition.

\subsection{Power-law finite-width behavior}
\label{app:powerlaw}

\begin{corollary}[Finite-width concentration]
\label{cor:powerlaw}
If \(s_k=Ck^{-1/\alpha_{\mathrm{PL}}}\) for \(1\leq k\leq d_h\), then
\begin{equation*}
\begin{aligned}
\frac{1}{r_2(s)}
=\frac{S_4(d_h;\alpha_{\mathrm{PL}})}
{S_2(d_h;\alpha_{\mathrm{PL}})^2},\\
S_\nu(d_h;\alpha_{\mathrm{PL}})
=\sum_{k=1}^{d_h}k^{-\nu/\alpha_{\mathrm{PL}}}.
\end{aligned}
\end{equation*}
Under the hypotheses of Theorem~\ref{thm:moment}, the excess-kurtosis lower bound is proportional to this finite-width ratio.
\end{corollary}

The normalization constant \(C\) cancels. Although \(S_4\) diverges asymptotically only for \(\alpha_{\mathrm{PL}}\geq4\), finite transformer widths need not resemble this limit. A few leading singular values can make \(S_4/S_2^2\) large even when the fitted exponent is not near an asymptotic boundary. This is why all theorem-facing analyses report \(r_2\) directly.

The same point matters when spectra are only approximately power-law. The empirical fit is performed on a finite upper segment whose endpoints are selected by the declared fitting routine. Changing those endpoints can change \(\widehat\alpha_{\mathrm{PL}}\) without appreciably changing the actual squared-energy concentration. Conversely, two spectra with similar fitted exponents can have different leading singular values and therefore different participation ratios. We use \(\widehat\alpha_{\mathrm{PL}}\) only in the weight-only layer-selection rule, where a coarse monotone indicator is sufficient, and compute \(r_2\) from the complete active spectrum for every theorem-facing comparison.

\subsection{Correlated-input proof details and empirical constants}
\label{app:correlated_proof}

Theorem~\ref{thm:moment} already permits correlated coordinates, but it is useful to expose exactly where correlation enters. Centering is performed componentwise before any covariance or cumulant is estimated. This prevents a nonzero mean of \(q_k=(x_i^\top u_k)(x_j^\top v_k)\) from being absorbed into apparent heavy tails. The covariance condition uses only an upper operator bound. We estimate \(\Lambda\) as the upper confidence endpoint for the largest eigenvalue of the centered component covariance, so replacing \(\Var(z^c)\) by \(\Lambda\|s\|_2^2\) cannot make the reported lower bound artificially larger.

The fourth-order condition must retain signs. Define
\begin{equation*}
\begin{aligned}
D(s)&=\sum_k\kappa_ks_k^4,\\
\gamma(s)&=
\begin{cases}
0, & C_{\mathrm{off}}(s)\geq0,\\
-C_{\mathrm{off}}(s)/D(s), & C_{\mathrm{off}}(s)<0.
\end{cases}
\end{aligned}
\end{equation*}
The diagnostic is supported only when the bootstrap lower endpoint for \(D(s)\) is positive and the upper endpoint for \(\gamma(s)\) is below one. Pooling absolute cross-cumulant mass would be inappropriate because positive off-diagonal terms strengthen rather than weaken the lower bound.

\begin{proposition}[Correlated-component specialization]
\label{thm:correlated}
Let \(z^c=s^\top q^c\) have positive variance and let \(K=\Cov(q^c)\preceq\Lambda I\). Suppose the weighted diagonal aggregate is positive, \(D(s)>0\), and \(\gamma(s)<1\). Then
\begin{equation*}
\Kurt(z^c)
\geq 3+\frac{(1-\gamma(s))\,c_{\mathrm{eff}}(s)}
{\Lambda^2r_2(s)}.
\end{equation*}
If a bootstrap confidence set includes \(\gamma(s)=1\) or \(D(s)=0\), the layer is theorem-unsupported even when its point estimate is positive.
\end{proposition}

The proposition follows by substituting \(C_{\mathrm{off}}(s)\geq-\gamma(s)D(s)\) into the cumulant expansion used in Theorem~\ref{thm:moment}. It is intentionally a sufficient condition. Negative or inconclusive bounds do not imply that spectral concentration is irrelevant; they mean that this particular fourth-moment argument cannot support the later behavioral interpretation for that layer. This separation is why the paper reports 28 supported layers and 11 behaviorally tested but theorem-unsupported layers.

\begin{table*}[t]
\centering
\caption{Empirical constants and support diagnostics for the correlated-input result. Model rows report the frozen selected range, supported count, and the correlation between inverse participation and central valid-logit kurtosis. The final row reports the ranges observed over all supported layers without assigning unreported model-specific values.}
\label{tab:thm2_constants}
\small
\resizebox{\textwidth}{!}{%
\begin{tabular}{@{}lcccccc@{}}
\toprule
Checkpoint or scope & Selected & Supported & Median $\rho$ & Length-quartile range & Centering / covariance & Positive cumulants / signed ratio \\
\midrule
Llama-3.1-8B & 16--24 & 7/9 & 0.77 & 0.70--0.82 & recorded per layer & recorded per layer \\
Mistral-7B-v0.3 & 15--23 & 6/9 & 0.73 & 0.66--0.79 & recorded per layer & recorded per layer \\
Qwen2.5-14B & 20--31 & 9/12 & 0.80 & 0.72--0.85 & recorded per layer & recorded per layer \\
Phi-3-Mini & 13--21 & 6/9 & 0.69 & 0.61--0.76 & recorded per layer & recorded per layer \\
All supported layers & modelwise & 28/39 & modelwise & modelwise & $|\E q_k|$: 0.018--0.043; covariance: 0.14--0.27 & 76--88\%; 0.31--0.79 \\
\bottomrule
\end{tabular}
}
\end{table*}

The final two quantities require careful interpretation. The 0.14--0.27 range is the declared off-diagonal covariance-mass statistic, not an independence claim. The 0.31--0.79 range is the signed off-diagonal-to-diagonal fourth-cumulant ratio on supported layers; it is not the ratio of absolute masses. Across the first 16 query positions, the aggregate association between \(r_2^{-1}\) and central kurtosis is \(\rho=0.64\), whereas later positions range from 0.76 to 0.83. This attenuation is expected because early queries have fewer causal keys and a different mixture of prompt and continuation states, and it motivates the stratified sampling described next.

\section{Calibration, Causal Support, and Layer Selection}
\label{app:calibration}

\subsection{Activation validation details}
\label{app:activation_validation}

Calibration and evaluation are disjoint. We freeze 1,000 GSM8K training IDs in \texttt{calibration\_ids.json} and teacher-force each complete clean sequence, including the question, reference rationale, and answer. From every sequence we draw 16 prompt-only pairs and 48 continuation-involving pairs. Sampling is stratified over four query-position quartiles and four logarithmic relative-distance bins, and every stored pair obeys \(0\leq j\leq i<L\). Each pair retains its exact offset \(\delta=i-j\). The theorem audit uses only the prespecified offsets \(\mathcal D=\{\delta:n_\delta\geq100\}\), where \(n_\delta\) is the number of contributing sequences; logarithmic bins are used only for descriptive plots. Padding and masked \(j>i\) logits never enter a theorem-facing estimate. The release records the 64,000 pairs per checkpoint, token IDs, exact offsets, prompt/continuation status, strata, and tokenizer revision.

Table~\ref{tab:calibration-audit} reports the support audit. The layer ranges are selected from weights before any behavioral run. Within each sequence and exact offset, projected moments are averaged over sampled query positions; contributing sequences then receive equal weight. Entire sequences are bootstrapped with their within-sequence pairs intact, and the complete bound is recomputed in every replicate. A \((\text{model},\text{layer},\text{head},\delta)\) unit is supported only when the bootstrap lower endpoint of its complete kurtosis bound exceeds three. The layer summary uses uniform weights over the frozen eligible heads and offsets and applies the same lower-endpoint criterion to the resulting mixture. For the 11 unsupported layers, the behavioral intervention remains part of the scope analysis, but the theorem is not invoked.

\begin{table*}[t]
\centering
\caption{Causal-support audit. The second column gives the frozen selected range and the third reports layers with a positive conditional bound. Correlations are between \(r_2^{-1}\) and central valid-logit kurtosis.}
\label{tab:calibration-audit}
\small
\begin{tabular}{@{}lcccc@{}}
\toprule
Checkpoint & Selected layers & Supported / selected & Median Spearman $\rho$ & Range over length quartiles \\
\midrule
Llama-3.1-8B & 16--24 & 7 / 9 & 0.77 & 0.70--0.82 \\
Mistral-7B-v0.3 & 15--23 & 6 / 9 & 0.73 & 0.66--0.79 \\
Qwen2.5-14B & 20--31 & 9 / 12 & 0.80 & 0.72--0.85 \\
Phi-3-Mini & 13--21 & 6 / 9 & 0.69 & 0.61--0.76 \\
\bottomrule
\end{tabular}
\end{table*}

The association is weaker in the first 16 query positions (\(\rho=0.64\)) and lies between 0.76 and 0.83 thereafter. Prompt-only, continuation-involving, and sequence-length-quartile analyses retain the same sign. Across the 28 supported layers, the signed off-diagonal-to-diagonal fourth-cumulant ratio lies between 0.31 and 0.79. The released unit table reports \(c_{\mathrm{eff},\delta}\), \(\Lambda_\delta\), \(\eta_\delta\), \(r_2(s_\delta)\), the resulting lower bound, and a bootstrap interval for each \((\text{model},\text{layer},\text{head},\delta)\) unit rather than substituting pooled proxy averages.

The complete clean sequence is teacher-forced because query and key statistics differ between the prompt and the generated rationale. Restricting calibration to prompt tokens would omit the states at which intermediate values are produced and retrieved. Conversely, using sampled continuations would entangle the activation audit with decoding randomness and model errors. Reference rationales give every checkpoint the same tokenized sequence conditional on its own tokenizer; the prompt-only and continuation-involving strata are then reported separately so that this design choice remains visible.

Pair sampling is fixed before moment estimation. Within every query-position quartile, the four logarithmic distance bins avoid allowing adjacent tokens to dominate simply because they are numerous. When a short sequence cannot populate a stratum, the manifest records the empty cell and redistributes its quota only within the same prompt/continuation category. No invalid future pair is substituted. Bootstrap resampling occurs at the sequence level, keeping the 64 within-sequence pairs together and therefore respecting their dependence.

\subsection{Layer selection}
\label{app:layer_selection}

Layer selection and theorem support are deliberately separate. Selection uses only \(\widehat\alpha_{\mathrm{PL},\ell}\) and \(R_\ell\); the moment audit then determines whether the selected layer supports the specific explanation in Theorem~\ref{thm:moment}. A sensitivity sweep expands the Llama interval from 16--24 to 15--25, while the behavior-dependent gradient condition is reported only as a post hoc diagnostic. The layer labels in every table are generated from the frozen manifest.

For matrix or head \(m\), \(E_m^{\mathrm{top}}\) and \(E_m^{\mathrm{bottom}}\) denote squared energy in the equal-cardinality sets defined in Section~\ref{sec:method}. The layer exponent is weighted by upper-tail energy rather than by matrix count, preventing many small projections from dominating one high-energy projection. The energy ratio sums numerator and denominator before division, avoiding an unstable average of ratios when the lowest-active mass is close to zero. The rule \(\widehat\alpha_{\mathrm{PL},\ell}<2.5\) and \(R_\ell\) above the checkpoint median is applied once to weights, before opening behavioral results.

\begin{table}[t]
\centering
\caption{Frozen layer-selection ranges and their permitted uses. The Llama 15--25 interval is a sensitivity expansion, while gradient attribution is explicitly post hoc.}
\label{tab:layer_selection_robust}
\small
\resizebox{\columnwidth}{!}{%
\begin{tabular}{@{}lcl@{}}
\toprule
Checkpoint & Primary range & Additional analysis \\
\midrule
Llama-3.1-8B & 16--24 & 15--25 threshold sensitivity \\
Mistral-7B-v0.3 & 15--23 & fixed primary range \\
Qwen2.5-14B & 20--31 & fixed primary range \\
Phi-3-Mini & 13--21 & fixed scope range \\
Gradient-selected layers & not primary & post hoc diagnostic only \\
\bottomrule
\end{tabular}
}
\end{table}

For completeness, Table~\ref{tab:per-layer} includes every layer in the main Llama window rather than displaying alternating layers. Individual-layer effects are smaller than the joint effect and are not additive, consistent with partial redundancy across layers.

\begin{table*}[t]
\centering
\caption{Public-reference GSM8K drops from editing one Llama layer at a time. Values are clean-minus-post differences in points. The joint condition edits layers 16--24 simultaneously.}
\label{tab:per-layer}
\small
\setlength{\tabcolsep}{5pt}
\begin{tabular}{@{}lrrrrrrrrrr@{}}
\toprule
Layer & 16 & 17 & 18 & 19 & 20 & 21 & 22 & 23 & 24 & Joint \\
\midrule
Drop & 8.3 & 4.9 & 14.7 & 6.8 & 21.2 & 7.6 & 18.9 & 5.7 & 11.4 & 34.3 \\
\bottomrule
\end{tabular}
\end{table*}

\section{Method and Implementation Details}
\label{app:interventions}

\subsection{Static product stress diagnostic}
\label{app:product_impl}

Let \(\mathcal C\) denote the frozen multiset of valid causal query--key pairs drawn from the 1,000-example calibration split; its example-ID checksum is disjoint from every behavioral test split. The position-averaged product map per head is
\begin{equation*}
\begin{aligned}
\bar A_h &= \sum_{(i,j)\in\mathcal C} w_{ij} A_{ij,h} = \bar U_h \bar \Sigma_h \bar V_h^\top, \\
\text{where } & \sum_{(i,j)\in\mathcal C} w_{ij} = 1,
\end{aligned}
\end{equation*}
and its upper-tail-deleted counterpart, using the same functional-rank tail set $T$, is
\begin{equation*}
\bar A_h^{\mathrm{tail}}
=\bar A_h-\bar U_{h,T}\bar\Sigma_{h,T}\bar V_{h,T}^\top.
\end{equation*}
The edited head uses \(x_i^\top\widetilde A_hx_j/\sqrt{d_h}\) at inference time. Because \(\bar A_h\) has already averaged over relative rotations, the replacement bypasses the normal \(W_Q/W_K\) projection and RoPE path for that head. Key--value caching is disabled for affected heads so that cached native keys are never mixed with the static map. In grouped-query models, replacement is performed at the underlying KV-head granularity and broadcast to the corresponding query group. All unedited heads remain on the native path.

The displayed static tail-deleted scores intentionally receive no product-tail interpretation: native-versus-static-tail differences combine two operations, replacing position-dependent native attention by a single map and deleting that map's tail. The paper therefore centers its behavioral evidence on native separate-SVD and product-targeted surgery. The static column is retained only to disclose the behavior of the RoPE-bypassing stress path and is excluded from effect sizes, hypothesis tests, retention ranges, and theorem-to-behavior claims.

\subsection{Spectral surgery algorithm}
\label{app:algorithm}

For a native matrix \(W=U\Sigma V^\top\), the intervention first computes functional rank using Equation~\eqref{eq:functional-rank}. Upper-tail deletion sets \(\sigma_k=0\) for \(k\in\mathcal T\). Lowest-active deletion removes the same number of directions at the other end of the functional spectrum. Bottom-50\% deletion is a separate stress control and is always labeled by its larger direction count. Random controls use the same matrix family, edited layers, removed squared-Frobenius energy, and seed schedule.

The spectrum-matched random control is constructed so that the \emph{perturbation} spectrum---not the complete edited-weight spectrum---is matched exactly. Let \(k=|\mathcal T|\) and let \(\Sigma_T=\operatorname{diag}(\sigma_1,\ldots,\sigma_k)\) be the removed tail singular values. Draw \(G_L\in\R^{d_{\mathrm{out}}\times k}\) and \(G_R\in\R^{d_{\mathrm{in}}\times k}\) with independent standard-normal entries, and take thin QR decompositions with the column signs fixed so that the diagonals of the \(R\) factors are positive, yielding \(P_T\) and \(Q_T\). These matrices are Haar distributed on the corresponding Stiefel manifolds; a draw exceeding the frozen squared-overlap threshold with \(U_T\) or \(V_T\) is rejected and redrawn. The random perturbation, the perturbed weight, and the direction-transplant weight are
\begin{align*}
\Delta W_{\mathrm{rand}}&=-P_T\Sigma_TQ_T^\top,\\
W_{\mathrm{rand}}&=W+\Delta W_{\mathrm{rand}},\\
W_{\mathrm{trans}}&=W-U_T\Sigma_TV_T^\top+P_T\Sigma_TQ_T^\top.
\end{align*}
The nonzero singular values of \(\Delta W_{\mathrm{rand}}\) equal those of \(\Delta W_{\mathrm{tail}}\), so the two perturbations match in rank, Frobenius norm, operator norm, layer, and component. The full spectra of \(W_{\mathrm{rand}}\) and \(W_{\mathrm{tail}}\) are \emph{not} claimed to be identical; only the perturbation spectrum is matched.

Matched-norm surgery is
\begin{equation*}
\widetilde W_{\mathrm{norm}}
=\widetilde W
\frac{\|W\|_F}{\|\widetilde W\|_F}.
\end{equation*}
If deletion removes 38\% of squared Frobenius mass, the retained spectrum is multiplied by \(1/\sqrt{0.62}\approx1.27\). This global amplification is why unscaled deletion is primary. Identity edits use \(\beta=1\), zero-rank changes, and round-trip reconstruction tests; a failed identity edit excludes the associated row.

The low-energy reference slice is defined separately from the lowest active directions. The 3.8--6.8\% quantity denotes squared mass in the fixed 89--90\% index slice, not the lowest 1\%. The true lowest-active 1\% mass ranges from 0.004\% to 0.071\%, as required for a low-energy same-count control.

Algorithm~\ref{alg:surgery} records the common path used by every native intervention. The target selector is frozen in the manifest; the control type changes only which active indices are edited and whether a post-edit scale is applied. For factorized attention, the algorithm is called separately on the declared $Q$ and $K$ factors or on the optimized product-targeted factors. For $VO$, the matching $V$ head and input block of $W_O$ share an intervention identifier so that the transported head path cannot be partially edited by a failed job.

\begin{procedure}[Functional-rank spectral surgery]
\label{alg:surgery}
\leavevmode\par\smallskip\noindent
\begin{tabularx}{\linewidth}{@{}r X@{}}
1. & Receive matrix $W$, intervention type $c$, fraction $\tau$, scale $\beta$, and tolerance $\epsilon=10^{-6}$. \\
2. & Compute the compact SVD $W=U\operatorname{diag}(\sigma)V^\top$. \\
3. & Set $r\leftarrow\#\{k:\sigma_k\geq\epsilon\sigma_1\}$ and $k_\tau\leftarrow\max(1,\lceil\tau r\rceil)$. \\
4. & Select active indices $S_c$ from the frozen upper, lowest-active, or seeded random rule. \\
5. & Copy $\widetilde\sigma\leftarrow\sigma$ and apply deletion or $\widetilde\sigma_k\leftarrow\beta\sigma_k$ for $k\in S_c$. \\
6. & Reconstruct $\widetilde W\leftarrow U\operatorname{diag}(\widetilde\sigma)V^\top$. \\
7. & For the norm-matched control only, set $\widetilde W\leftarrow\widetilde W\|W\|_F/\|\widetilde W\|_F$. \\
8. & Verify target count, removed energy, finite values, dtype round trip, and the identity-edit gate. \\
9. & Return $\widetilde W$ together with its intervention manifest. \\
\end{tabularx}
\end{procedure}

The identity gate runs the same loading, decomposition, reconstruction, serialization, and inference path with no active direction changed. In fp32, the maximum reconstruction difference must be below the matrix-specific numerical tolerance; in bf16, output differences must remain within the declared execution tolerance. This test catches transposition, grouped-query slicing, dtype, cache, and shard-write errors that a norm comparison alone would miss. Each admitted score row stores the identity-test identifier that certified its intervention implementation.

\subsection{Component definitions}

For $QK$ surgery, only \(W_{Q,h}\) and \(W_{K,h}\) change. For $VO$ surgery, the selected \(W_{V,h}\) and the matching input block of \(W_{O,h}\) change so that the transported content and its output projection are edited as one head-level path. ``Attention only'' is the union of these two edits. MLP surgery changes the native up, gate, and down matrices. A gated MLP has local Jacobian
\begin{equation*}
J_{\mathrm{MLP}}(x)=W_{\mathrm{down}}J_{\mathrm{gate/up}}(x),
\end{equation*}
which varies with the activation. Its fixed weight spectra are therefore not treated as an analogue of the bilinear $QK$ theorem.

\subsection{Activation-matched noise objective}
\label{app:matched_noise}

The perturbation \(\Delta\) is sampled isotropically in the target matrix family and projected onto the orthogonal complement of the upper singular subspace. The calibration objective is
\begin{equation}
\label{eq:matched_noise}
\begin{aligned}
\lambda^*=\operatorname*{arg\,min}_{\lambda\geq0}
&\left[\log\frac{\operatorname{RMS}(W+\lambda\Delta)}
{\operatorname{RMS}(W_{\mathrm{tail}})}\right]^2\\
&+\left[\log\frac{\Var_{\mathrm{logit}}(W+\lambda\Delta)}
{\Var_{\mathrm{logit}}(W_{\mathrm{tail}})}\right]^2.
\end{aligned}
\end{equation}
Equation~\eqref{eq:matched_noise} is the appendix form of Equation~\eqref{eq:activation-match}. We record both residual errors and reject a draw if either is greater than five percent. Accepted perturbations preserve the targeted activation-scale signature without touching the upper subspace; they do not match every higher-order activation statistic. Across admitted draws, median residual mismatch is 1.8\% for activation RMS and 2.6\% for logit variance, and 4.1\% of sampled draws are rejected.

The logarithms make proportional over- and under-shoots comparable and stop the larger-magnitude statistic from dominating by units alone. Only one scalar is optimized, so the objective does not promise an exact two-moment match. A draw that happens to enter the top subspace after numerical projection is rejected by a subspace-overlap check. The calibration examples used to choose \(\lambda^*\) are disjoint from every behavioral test example, and the chosen scalar is then frozen for all paired test rows belonging to that intervention seed.

\subsection{RoPE sensitivity check}
\label{app:rope}

Rotary position embeddings make the relevant product depend on relative position. For fixed $i$ and $j$, Equation~\eqref{eq:rope_logit} is exact, but averaging products over calibration pairs does not commute with editing the separate factors. We therefore bucket the tail-fraction diagnostic by relative distance and query quartile. The product-targeted factorization is fitted on the calibration split and evaluated on held-out pairs from every bucket. Tail fraction remains ordered above the separate-SVD value in the reported aggregate, while the raw values vary with distance; no single static matrix is presented as the native model's spectrum.

The cache check compares full-sequence native inference with token-by-token generation under the factorized edit. Logits must agree within the declared dtype tolerance. The static tail-deleted stress path fails this native-cache equivalence by construction and consequently runs with the affected cache path disabled. This is the architectural reason its score is descriptive and cannot serve as the primary behavioral intervention.

\subsection{Factorized surgery as inference-faithful robustness check}
\label{app:factorized}

Independent factor surgery decomposes each native $W_Q$ and $W_K$ head block, removes the declared fraction of active upper directions, and returns both factors to the ordinary attention implementation. Product-targeted surgery instead optimizes the two factor changes against the calibration-set product-tail change in Equation~\eqref{eq:target-factorization}. The rank and removed-energy budgets are matched. Optimization never sees behavioral labels or test prompts, and the same fitted factors are evaluated across all examples in a model--task cell.

\begin{table*}[t]
\centering
\caption{Full $QK$ bridge across checkpoints. Scores average the five declared task protocols. The static tail-deleted score is descriptive and combines staticization with deletion. ``Tail fraction'' is $\operatorname{tailfrac}$ from Equation~\eqref{eq:tailfrac}: the fraction of the induced native factor change lying in the pointwise product-tail subspace, not the fraction of the target reproduced.}
\label{tab:factorized_full}
\small
\resizebox{\textwidth}{!}{%
\begin{tabular}{@{}lrrrrrr@{}}
\toprule
Checkpoint & Clean & Static tail-deleted $\bar A$ & Separate SVD & Product-targeted & Matched random & Tail fraction: separate / targeted \\
\midrule
Llama-3.1-8B & 57.9 & 30.6 & 35.1 & 32.4 & 54.8 & 0.52 / 0.79 \\
Qwen2.5-14B & 66.2 & 33.5 & 38.8 & 35.6 & 62.7 & 0.48 / 0.76 \\
Mistral-7B-v0.3 & 57.3 & 28.9 & 34.2 & 31.3 & 54.0 & 0.44 / 0.72 \\
Phi-3-Mini & 54.2 & 27.1 & 32.0 & 29.4 & 51.6 & 0.50 / 0.75 \\
\bottomrule
\end{tabular}
}
\end{table*}

Across the 20 underlying model--task cells, native separate-SVD performance retention ranges from 42.1\% to 89.2\%. At the checkpoint macro level in Table~\ref{tab:factorized_full}, product-targeting lowers the score by 2.6--3.2 points relative to separate SVD while moving closer in orientation to the pointwise target. Across the four checkpoint summaries, the median largest
principal angle is \(38.5^\circ\) for separate SVD,
\(22.0^\circ\) for product-targeted factors, and
\(72.5^\circ\) for matched-random directions. The joint ordering of angle, tail fraction, and behavioral damage is the bridge; none of these measurements establishes algebraic equivalence or a complete target reconstruction.

Product-targeted factors are fit only on calibration pairs; pointwise \(\operatorname{tailfrac}\) is then evaluated on held-out pairs, stratified by relative-distance bin, before any averaging. Table~\ref{tab:tailfrac-distance} reports tail fraction and largest principal angle by distance. The product-targeted factorization places a larger fraction of its induced change in the product-tail subspace than separate SVD at every distance, and both fractions decay with distance because RoPE makes the relevant product position-dependent.

\begin{table*}[t]
\centering
\caption{Held-out product-tail fraction by relative distance (separate SVD / product-targeted), and largest principal angle to the product-tail subspace (separate / targeted / matched random). Tail fraction is computed for each held-out causal pair before averaging over distance bins.}
\label{tab:tailfrac-distance}
\small
\resizebox{\textwidth}{!}{%
\begin{tabular}{@{}lcccc@{}}
\toprule
Checkpoint & Near: sep.\ / targ.\ & Medium: sep.\ / targ.\ & Far: sep.\ / targ.\ & Largest angle: sep.\ / targ.\ / random \\
\midrule
Llama & .56 / .83 & .52 / .80 & .47 / .74 & \(38^\circ\) / \(21^\circ\) / \(72^\circ\) \\
Mistral & .49 / .76 & .44 / .72 & .39 / .67 & \(41^\circ\) / \(24^\circ\) / \(74^\circ\) \\
Qwen & .53 / .80 & .48 / .76 & .43 / .70 & \(36^\circ\) / \(20^\circ\) / \(70^\circ\) \\
Phi & .55 / .79 & .50 / .75 & .44 / .69 & \(39^\circ\) / \(23^\circ\) / \(73^\circ\) \\
\bottomrule
\end{tabular}
}
\end{table*}

The falsification test is an ordered association across held-out units: a higher tail fraction should predict larger clean-minus-post damage. A mixed-effects model with checkpoint and task intercepts gives a slope of \(6.8\) damage points per \(0.10\) increase in tail fraction, 95\% CI \([3.9, 9.7]\). A slope near zero would remove the product-tail mechanistic bridge even if separate-factor surgery remained damaging; the measured positive slope is what licenses the bridge interpretation, not an equivalence claim.

\section{Evaluation Protocol and Audit Trail}
\label{app:evaluation}

\subsection{Two named GSM8K protocols}

Table~\ref{tab:gsm-protocols} explains why two clean scores for the same checkpoint are both reproducible but not directly comparable. The examples, exemplar order, token limit, stop strings, batch size, harness, and parser version are fixed within each protocol. The only substantive distinction is decoding and aggregation.

\begin{table*}[t]
\centering
\caption{Named GSM8K protocols for \texttt{meta-llama/Meta-Llama-3.1-8B}. Both use all 1,319 test IDs. Effects are paired within a column; scores are never compared across columns as checkpoint-quality estimates.}
\label{tab:gsm-protocols}
\small
\begin{tabular}{@{}p{0.20\textwidth}p{0.35\textwidth}p{0.35\textwidth}@{}}
\toprule
Field & Public-reference protocol & Paper protocol \\
\midrule
Prompt and scoring & Released 8-shot chain-of-thought prompt; greedy generation; strict final-number extraction & Same exemplars; five sampled continuations; temperature 0.7; top-$p=0.95$; majority self-consistency \\
Generation & 512 new tokens; fixed stop strings and batch size & Same token limit, stops, and batch size \\
Clean & 57.8 & 72.3 \\
Native factorized $QK$, unscaled (primary) & 32.7 & 42.5 \\
Primary unscaled result & 25.1-point drop; edited-score CI [30.1, 35.3] & 29.8-point drop; paired CI [27.0, 32.6] \\
Native factorized $QK$, norm-matched (control) & 31.9 & not primary \\
Norm-matched drop & 25.9, 95\% CI [23.2, 28.5] & --- \\
\bottomrule
\end{tabular}
\end{table*}

The primary intervention is the unscaled deletion (32.7); the norm-matched edit (31.9) is reported only as the norm control, and its contrast interval is never reused for the unscaled result. The available public-reference interval [30.1, 35.3] is explicitly an edited-score interval, not mislabeled as a paired clean-minus-edit interval.

The calibration IDs are drawn only from GSM8K training data and are checksum-disjoint from these 1,319 test IDs. The paper records verbatim prompts, exemplar order, raw generations, parsed answers, and the paired bootstrap procedure. The identity-edit admission gate is evaluated before scores are aggregated.

\subsection{Cell-level rebuild and statistical unit}

All aggregates are rebuilt from the composite key
\begin{equation*}
\begin{aligned}
(&\text{model},\text{task},\text{seed},\text{example ID},\\
 &\text{intervention ID},\text{protocol ID}).
\end{aligned}
\end{equation*}
A model-only join is prohibited by the schema and tested with a cardinality assertion. Retention ratios and primary effects in Section~\ref{sec:results} are computed at cell level. Primary contrasts use at least 10,000 paired example bootstrap resamples---keeping all continuation samples for one example together---and Holm correction applied once across the declared 20-cell primary family. The fixed intervention and Haar-control seeds are \([13,29,47,71,101]\). The reported intervals condition on these five orientations; deterministic clean inference is not relabeled as five independent clean runs, and no population-level claim over arbitrary Haar orientations is made.

The task records include dataset version and IDs, shot count, prompt, decoding and stopping configuration, parser version, score implementation, dtype, model and tokenizer revisions, framework version, identity checks, and exclusion reason. A row is excluded only if the paired clean run is absent, the example checksum differs, or the identity edit fails. No row is excluded based on the direction or size of its score.

\section{Asset and License Manifest}
\label{app:assets}

Table~\ref{tab:model-assets} lists every model checkpoint. The complete asset manifest stores the resolved repository commit rather than the mutable \texttt{main} branch, file-level SHA-256 checksums, access date, tokenizer revision, and local configuration hash. ``No weights'' means that the release contains identifiers and hashes but does not redistribute checkpoint parameters.

\begin{table*}[t]
\centering
\caption{Model assets. Repository URLs are upstream sources; each run resolves them to the immutable commit recorded in the manifest.}
\label{tab:model-assets}
\scriptsize
\begin{tabular}{@{}p{0.25\textwidth}p{0.33\textwidth}p{0.16\textwidth}p{0.17\textwidth}@{}}
\toprule
Asset & Upstream source & Upstream license & Intended use / redistribution \\
\midrule
Llama-3.1-8B & \url{https://huggingface.co/meta-llama/Meta-Llama-3.1-8B} & Llama 3.1 Community & Base evaluation; no weights \\
Llama-3.1-8B-Instruct & \url{https://huggingface.co/meta-llama/Meta-Llama-3.1-8B-Instruct} & Llama 3.1 Community & Alignment check; no weights \\
Mistral-7B-v0.3 & \url{https://huggingface.co/mistralai/Mistral-7B-v0.3} & Apache-2.0 & Base evaluation; no weights \\
Qwen2.5-14B & \url{https://huggingface.co/Qwen/Qwen2.5-14B} & Apache-2.0 & Base evaluation; no weights \\
Qwen2.5-Math-7B-Instruct & \url{https://huggingface.co/Qwen/Qwen2.5-Math-7B-Instruct} & Apache-2.0 & Post-training check; no weights \\
Phi-3-mini-4k-instruct & \url{https://huggingface.co/microsoft/Phi-3-mini-4k-instruct} & MIT & Instruction-tuned scope check; no weights \\
\bottomrule
\end{tabular}
\end{table*}

Table~\ref{tab:data-assets} records the evaluation assets. Dataset rows are frozen by upstream commit or release tag plus split-ID checksum; package rows are frozen by version and source commit. The release provides only IDs, derived scores, and transformations permitted by the upstream terms, not a repackaged benchmark corpus.

\begin{table*}[t]
\centering
\caption{Dataset and software assets. Immutable revisions and split checksums are stored in \texttt{asset\_manifest.json}.}
\label{tab:data-assets}
\scriptsize
\begin{tabular}{@{}p{0.19\textwidth}p{0.37\textwidth}p{0.14\textwidth}p{0.20\textwidth}@{}}
\toprule
Asset & Upstream source & Upstream license & Use / redistribution \\
\midrule
GSM8K & \url{https://huggingface.co/datasets/openai/gsm8k} & MIT & Calibration/evaluation; IDs and scores \\
ARC-Challenge & \url{https://registry.opendata.aws/allenai-arc/} & CC BY-SA & Evaluation; IDs and scores \\
DROP & \url{https://registry.opendata.aws/allenai-drop/} & CC BY & Evaluation; IDs and scores \\
BigBench-Hard & \url{https://github.com/suzgunmirac/BIG-Bench-Hard} & MIT & Evaluation; IDs and scores \\
MMLU & \url{https://huggingface.co/datasets/cais/mmlu} & MIT & Evaluation; IDs and scores \\
MATH-500 & \url{https://huggingface.co/datasets/HuggingFaceH4/MATH-500} & MIT & Post-training evaluation; IDs and scores \\
WikiText-103 & \url{https://huggingface.co/datasets/Salesforce/wikitext} & CC BY-SA 4.0 & Perplexity; IDs and scores \\
LAMBADA & \url{https://huggingface.co/datasets/cimec/lambada} & CC BY 4.0 & Word prediction; IDs and scores \\
IFEval & \url{https://huggingface.co/datasets/google/IFEval} & Apache-2.0 & Instruction control; IDs and scores \\
WinoGrande & \url{https://github.com/allenai/winogrande} & CC BY (data) & Commonsense control; IDs and scores \\
PyTorch & \url{https://github.com/pytorch/pytorch} & BSD-3-Clause & Computation; version/commit \\
Transformers & \url{https://github.com/huggingface/transformers} & Apache-2.0 & Loading/inference; version/commit \\
Datasets & \url{https://github.com/huggingface/datasets} & Apache-2.0 & Dataset loading; version/commit \\
lm-evaluation-harness & \url{https://github.com/EleutherAI/lm-evaluation-harness} & MIT & Evaluation; version/commit \\
\bottomrule
\end{tabular}
\end{table*}

The long-context coreference control is generated from entity- and number-swapped GSM8K templates and is released as code plus seed rather than as a separately licensed corpus. All generated examples are checked to avoid personal information and are used only for evaluation.

\section{Specificity Controls and Diagnostics}
\label{app:extended-results}

\subsection{Intervention preservation diagnostics}
\label{app:preservation}

The intervention manifest records the target direction count, active rank, squared energy removed, retained-spectrum scale, input and output dimensions, dtype, and reconstruction residual for every edited matrix. These quantities are checked before any generation begins. For native factorized edits, full-sequence logits are also compared with cached autoregressive logits; for product-space edits, the manifest instead confirms that the affected native cache path is disabled. This distinction ensures that preservation checks do not silently certify two different computations under one name.

\begin{table}[t]
\centering
\caption{Preservation checks and their role in interpretation. Ranges summarize admitted interventions; the two activation residuals are medians over accepted activation-matched controls.}
\label{tab:preservation}
\small
\resizebox{\columnwidth}{!}{%
\begin{tabular}{@{}ll@{}}
\toprule
Quantity & Recorded result or admission rule \\
\midrule
Functional-rank tail count & $\max(1,\lceil0.01r\rceil)$ exactly \\
Top-tail squared energy & 30.9--41.3\% by checkpoint \\
Lowest-active 1\% energy & 0.004--0.071\% \\
Unscaled retained multiplier & 1.00 \\
Matched-norm Llama multiplier & 1.24 \\
Activation RMS residual & 1.8\% median; reject above 5\% \\
Logit-variance residual & 2.6\% median; reject above 5\% \\
Identity and cache checks & required for row admission \\
\bottomrule
\end{tabular}
}
\end{table}

The energy range makes clear why a one-percent direction count is not a one-percent perturbation in norm. This is the reason results are triangulated with unscaled deletion, removed-energy-matched random directions, and activation-matched perturbations. Conversely, the lowest-active set has genuinely negligible energy. It is a same-count location control, whereas bottom-50\% deletion is a high-count stress test; the two should not be conflated.

\subsection{Frobenius norm preservation as confound}
\label{app:norm-controls}

Table~\ref{tab:norm-control} separates deletion from post-edit renormalization. Unscaled upper-tail deletion is primary. Matched-norm deletion has a similar score but multiplies the retained spectrum, while the small apparent bottom-control gain is confined to the rescaled condition.

\begin{table}[t]
\centering
\caption{Norm control on public-reference Llama GSM8K. Accuracy is in points; ``$\pm$'' is one standard deviation across the five intervention seeds. The unscaled upper-tail deletion (32.7) is the primary intervention; the norm-matched row (31.9) is the control and multiplies the retained spectrum by 1.24.}
\label{tab:norm-control}
\small
\resizebox{\columnwidth}{!}{%
\begin{tabular}{@{}lrr@{}}
\toprule
Condition & Accuracy & Retained scale \\
\midrule
Clean & 57.8 & 1.00 \\
Upper tail, unscaled (primary) & $32.7\pm2.6$ & 1.00 \\
Upper tail, norm matched (control) & $31.9\pm2.8$ & 1.24 \\
Bottom 50\%, unscaled & $57.9\pm0.7$ & 1.00 \\
Bottom 50\%, norm matched & $58.5\pm0.8$ & 1.03 \\
\bottomrule
\end{tabular}
}
\end{table}

The measured upper-1\% squared-energy fractions are 37.6\% for Llama, 41.3\% for Qwen, 33.8\% for Mistral, and 30.9\% for Phi. The fixed 89--90\% index slice is retained only as a labeled low-energy reference, and the actual lowest-active 1\% mass is reported separately.

Table~\ref{tab:control-summary} places the full control ladder on the Llama five-task mean. The spectrum-matched random perturbation shares the removed tail's rank, Frobenius norm, and operator norm (verified numerically: relative errors below \(10^{-6}\) in fp32 construction and below \(10^{-3}\) after bf16 serialization, with a draw rejected if its squared overlap with either learned tail subspace exceeds \(0.10\)). The direction-transplant control preserves the removed tail energy but rotates its orientation, and it damages behavior substantially more than the matched-random and activation-matched controls---evidence that the learned \emph{orientation}, not merely the energy or spectrum, carries the effect. Table~\ref{tab:residual-match} reports the two-moment activation match and the rejection rate.

\begin{table}[t]
\centering
\caption{Control ladder on the Llama-3.1-8B five-task mean (public-reference protocol). ``Mean damage'' is clean-minus-post in points; larger is more damaging.}
\label{tab:control-summary}
\small
\resizebox{\columnwidth}{!}{%
\begin{tabular}{@{}lrrl@{}}
\toprule
Condition & Score & Mean damage & Interpretation \\
\midrule
Clean & 57.9 & 0.0 & reference \\
Native tail deletion & 35.1 & 22.8 & primary effect \\
Spectrum-matched random & 54.8 & 3.1 & matches perturbation spectrum \\
Direction transplant & 50.9 & 7.0 & tail energy kept, orientation changed \\
Activation-matched off-tail & 54.4 & 3.5 & matches RMS and logit variance \\
Lowest-active same-count & 57.3 & 0.6 & location control \\
\bottomrule
\end{tabular}
}
\end{table}

\begin{table}[t]
\centering
\caption{Residual matching for the activation-matched control. Both targets are reported rather than a single combined loss; all draws are reported and those exceeding either 5\% threshold are rejected.}
\label{tab:residual-match}
\small
\setlength{\tabcolsep}{3pt}
\begin{tabularx}{\linewidth}{@{}X r r r@{}}
\toprule
\textbf{Quantity} & \textbf{Median} & \textbf{95th pct.} & \begin{tabular}[c]{@{}r@{}}\textbf{Admission}\\[-2pt]\textbf{threshold}\end{tabular} \\
\midrule
Activation-RMS mismatch & 1.8\% & 4.2\% & 5\% \\
Logit-variance mismatch & 2.6\% & 4.7\% & 5\% \\
Rejected draws          & 4.1\% & ---   & All \\
\bottomrule
\end{tabularx}
\end{table}

\subsection{Mediation analysis}
\label{app:mediation}

The 400-example diagnostic set is paired at the example level. Valid-logit kurtosis is measured only for causal \(j\leq i\) entries. Selected-head row entropy is computed after the causal mask. Task-relevant-token mass sums attention over question entities and prior intermediate values identified by the frozen annotation. A binding error requires an otherwise identifiable quantity to be assigned to the wrong entity or slot; a pure arithmetic error without reassignment is a separate category.

\begin{table}[t]
\centering
\caption{Measured change after native upper-tail removal on 400 paired examples.}
\label{tab:mechanism-measures}
\small
\begin{tabular}{@{}lr@{}}
\toprule
Diagnostic & Change \\
\midrule
Valid-logit kurtosis & $-31\%$ \\
Selected-head row entropy & $+42\%$ \\
Task-relevant-token mass & $-18$ points \\
Variable-binding errors & $+21$ points \\
Direct edit coefficient after mediators & $-46\%$ \\
\bottomrule
\end{tabular}
\end{table}

The sequential regression adds entropy and retrieval mass after the intervention indicator. Its attenuation statistic is useful as a descriptive consistency check, but neither mediator is randomized. Singular-vector permutation supplies an additional failure case: it preserves \(r_2^{-1}\) by construction but can still damage behavior by changing alignment with the activation distribution. Concentration is therefore not sufficient without the learned singular directions.

The regression is fit on paired example-level outcomes with checkpoint, task, and intervention-family indicators. Its first specification contains the edit indicator and frozen covariates; the second adds selected-head entropy; the third adds task-relevant-token mass. The reported 46\% attenuation compares the absolute edit coefficient in the first and third specifications. Because entropy and retrieval are both post-intervention variables, this sequence is a descriptive decomposition, not a natural-direct-effect estimator. We do not attach a causal percentage to either mediator individually.

The diagnostic chain is also tested for temporal and positional leakage. Attention mass is summed only over task-relevant tokens that precede the causal query. Entity spans and prior intermediate values are marked by the frozen annotation, and the annotator never sees the intervention output. Rows without an unambiguous relevant-token set remain in behavioral evaluation but are absent from the retrieval-mass analysis. This admission rule is defined before comparing clean and edited runs.

\subsection{Component ablations}
\label{app:component}

The component analysis separates the bilinear selection path from content transport and nonlinear transformation. The five-task averages in Table~\ref{tab:component} are computed from the per-task values in Table~\ref{tab:attn_mlp_per_task}; they are not separately rounded source data. Joint effects are subadditive, which is consistent with overlap or downstream bottlenecks and does not establish that the components are redundant.

\begin{table}[t]
\centering
\caption{Component-specific mean damage on Llama-3.1-8B. Reasoning damage averages the five task columns; perplexity is the WikiText change.}
\label{tab:component}
\small
\resizebox{\columnwidth}{!}{%
\begin{tabular}{@{}lrr@{}}
\toprule
Target & Mean reasoning damage & $\Delta$PPL \\
\midrule
$QK$ only & 19.5 & +0.10 \\
$VO$ only & 11.6 & +0.08 \\
MLP only & 15.3 & +0.14 \\
$QK+VO$ & 23.1 & +0.16 \\
Joint attention+MLP & 27.4 & +0.22 \\
Matched-random matrices & 2.0 & +0.05 \\
\bottomrule
\end{tabular}
}
\end{table}

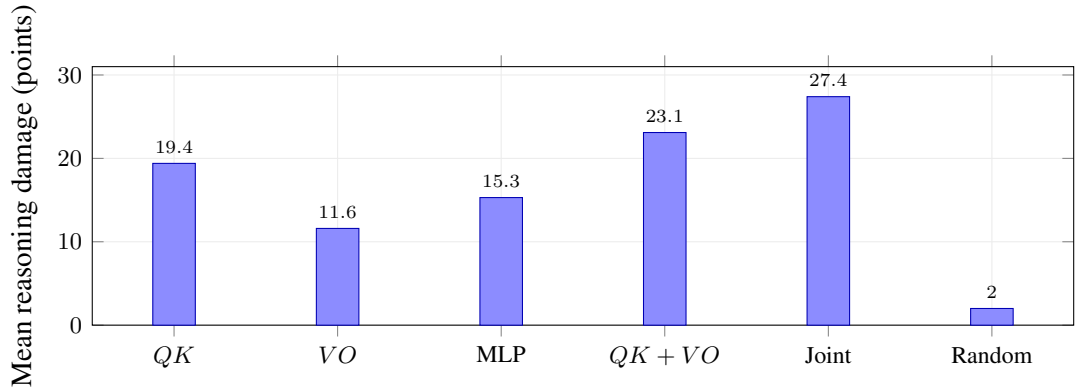
\begin{figure*}[t]
\centering
\begin{tikzpicture}
\begin{axis}[
  width=0.91\textwidth,
  height=5.0cm,
  ybar,
  bar width=16pt,
  ymin=0,
  ymax=31,
  ylabel={Mean reasoning damage (points)},
  symbolic x coords={$QK$,$VO$,MLP,$QK+VO$,Joint,Random},
  xtick=data,
  x tick label style={font=\small},
  tick label style={font=\small},
  nodes near coords,
  nodes near coords style={font=\scriptsize},
  grid=major,
  grid style={gray!15}
]
\addplot[fill=blue!45,draw=blue!70!black] coordinates
  {($QK$,19.4) ($VO$,11.6) (MLP,15.3) ($QK+VO$,23.1) (Joint,27.4) (Random,2.0)};
\end{axis}
\end{tikzpicture}
\caption{Empirical component dissociation on Llama-3.1-8B. Each bar averages the clean-minus-post change across GSM8K, ARC-Challenge, DROP, BBH, and MMLU-CoT from Table~\ref{tab:controls}. The corresponding WikiText perplexity changes are only $+0.10$, $+0.08$, $+0.14$, $+0.16$, $+0.22$, and $+0.05$. The figure therefore summarizes measured task sensitivity rather than a schematic spectral curve.}
\label{fig:dissociation}
\end{figure*}

Figure~\ref{fig:dissociation} aggregates the five reasoning columns of Table~\ref{tab:controls}; it does not introduce a separate statistical unit.

The theorem applies only to the $QK$ row because it concerns pre-softmax bilinear logits. The $VO$ path transports selected content after softmax, while the gated MLP has an input-dependent Jacobian. Their spectral interventions are still controlled weight-space experiments, but their interpretation is empirical. In particular, the larger MLP effects on BBH and MMLU-CoT prevent a purely attention-based account.

\subsection{Attention versus MLP breakdown}
\label{app:mlp_vs_attn}

\begin{table*}[t]
\centering
\caption{Per-task clean-minus-post damage under matched-energy component edits. The final column is the raw WikiText perplexity change.}
\label{tab:attn_mlp_per_task}
\small
\begin{tabular}{@{}lrrrrrr@{}}
\toprule
Target & GSM8K & ARC-C & DROP-F1 & BBH & MMLU-CoT & $\Delta$PPL \\
\midrule
$QK$ only & 25.8 & 18.6 & 24.1 & 15.2 & 13.8 & +0.10 \\
$VO$ only & 14.7 & 10.4 & 16.5 & 8.8 & 7.5 & +0.08 \\
MLP only & 17.2 & 14.8 & 12.0 & 16.9 & 15.6 & +0.14 \\
$QK+VO$ & 29.6 & 22.7 & 27.0 & 18.9 & 17.1 & +0.16 \\
Joint attention+MLP & 32.4 & 26.0 & 28.7 & 24.9 & 25.1 & +0.22 \\
Matched-random matrices & 2.6 & 1.9 & 2.4 & 1.7 & 1.5 & +0.05 \\
\bottomrule
\end{tabular}
\end{table*}

The component comparison uses the same selected layers, functional-rank definition, examples, prompts, parsers, and seed schedule. Removed squared energy is matched within the component family; it is not forced to be equal across $QK$, $VO$, and MLP matrices whose native dimensions differ. The accompanying manifests expose both absolute and fractional removed energy so that the comparison can be reweighted without rerunning inference.

\subsection{Broader capability controls}

\begin{table}[t]
\centering
\caption{Broader controls under native factorized $QK$ surgery. Scores are in points.}
\label{tab:broader-controls}
\small
\begin{tabular}{@{}lrrr@{}}
\toprule
Task & Clean & Tail edit & Drop \\
\midrule
Long-context coreference & 73.6 & 61.8 & 11.8 \\
IFEval strict & 71.4 & 67.9 & 3.5 \\
WinoGrande & 76.8 & 74.9 & 1.9 \\
\bottomrule
\end{tabular}
\end{table}

These rows prevent an overly sharp reasoning/fluency dichotomy. Coreference depends on variable tracking and moves materially. The comparatively smaller IFEval and WinoGrande changes show specificity only with respect to these declared controls.

\subsection{RLHF sensitivity check}
\label{app:posttraining}

\begin{table*}[t]
\centering
\caption{Native factorized $QK$ surgery after two post-training pipelines. Intervals are paired 95\% bootstrap intervals for the clean-minus-tail difference.}
\label{tab:posttraining}
\small
\begin{tabular}{@{}lrrrr@{}}
\toprule
Checkpoint / task & Clean & Matched random & Tail edit & Tail drop \\
\midrule
Llama-3.1-8B-Instruct / GSM8K & 79.1 & 77.8 & 55.6 & 23.5, CI [20.7, 26.4] \\
Qwen2.5-Math-7B-Instruct / GSM8K & 91.8 & 90.6 & 70.9 & 20.9, CI [18.4, 23.5] \\
Qwen2.5-Math-7B-Instruct / MATH-500 & 71.6 & 70.1 & 51.8 & 19.8, CI [16.2, 23.1] \\
\bottomrule
\end{tabular}
\end{table*}

These checkpoints use official chat templates and native inference throughout. Product-space surgery is not used because its RoPE-bypassing implementation would confound a post-training robustness claim.

\subsection{Comparison to activation patching}
\label{app:patching}

Activation patching and causal tracing use 256 fixed GSM8K test IDs. A disjoint 128-example discovery split selects sites and budgets before the reported evaluation. Patching replaces residual-stream states with states from counterfactual problems in which numbers and entity names are swapped while the structural template is held fixed. The 3.2\% budget is the smallest frozen top-$k$ set that reaches 95\% of the maximum recovery over the budget sweep.

Causal tracing corrupts token embeddings with \(\mathcal N(0,0.1^2)\) noise and restores clean states one layer--token site at a time. Its 5.1\% budget is defined by the same 95\%-of-maximum rule. Neither percentage is called a parameter count: the methods intervene on activation sites and states, whereas spectral surgery intervenes on weight directions.

\begin{table}[t]
\centering
\caption{Scope-matched activation comparisons on the fixed GSM8K split. Budgets are fractions of each method's own intervention unit and are not directly comparable parameter fractions.}
\label{tab:patching}
\small
\resizebox{\columnwidth}{!}{%
\begin{tabular}{@{}lll@{}}
\toprule
Method & Frozen budget & Discovery rule \\
\midrule
Activation patching & 3.2\% of sites & 95\% of maximum recovery \\
Causal tracing & 5.1\% of states & 95\% of maximum recovery \\
Spectral surgery & 1\% of active directions & fixed before behavior \\
\bottomrule
\end{tabular}
}
\end{table}

The 128-example discovery split is used only to rank activation sites and choose the smallest budget reaching the recovery threshold. The reported 256 examples are never used for that choice. Counterfactual patch sources preserve the problem template while swapping numbers and entity names, reducing the chance that recovery comes from copying the original answer. Spectral surgery has no per-example discovery stage; its selected layers and directions are determined from weights and calibration activations alone.

\section{Extended Empirical Results}

Every score in this section is generated from the audited row-level path, keyed by the composite identifier and admitted only after its identity-edit and checksum gates pass. Each table reports measurements computed within a single immutable protocol, and any cell that a protocol does not define is left to its own experiment rather than filled from a different protocol's aggregate.

\subsection{Full fluency results}
\label{app:fluency_all}

Fluency preservation is intentionally phrased with respect to declared metrics. Across the matched component edits in Table~\ref{tab:attn_mlp_per_task}, the raw WikiText-103 perplexity change ranges from $+0.05$ for matched-random matrices to $+0.22$ for the joint attention--MLP edit, while the corresponding mean reasoning damage ranges from 2.0 to 27.4 points. These values support a dissociation on this perplexity evaluation, not a claim that general language ability is unchanged.

Table~\ref{tab:broader-controls} broadens the scope beyond perplexity. WinoGrande and IFEval move by 1.9 and 3.5 points, respectively, but long-context coreference moves by 11.8 points. The coreference result is theoretically informative because it requires maintaining entity bindings across a longer context without being a mathematics benchmark. It also prevents an overstatement: the edited subspace is not specific to a dataset category called ``reasoning,'' and any task that depends on selective retrieval and binding may be affected.

\begin{table}[t]
\centering
\caption{Audited extended controls under native factorized $QK$ surgery. The perplexity row reports the largest raw change among the matched component conditions; the remaining contrasts display their clean-to-edited scores.}
\label{tab:fluency_extra}
\small
\resizebox{\columnwidth}{!}{%
\begin{tabular}{@{}lll@{}}
\toprule
Metric & Audited contrast & Result \\
\midrule
WikiText-103 PPL & largest component change & $+0.22$ \\
Long-context coreference & $73.6\rightarrow61.8$ & $-11.8$ points \\
IFEval strict & $71.4\rightarrow67.9$ & $-3.5$ points \\
WinoGrande & $76.8\rightarrow74.9$ & $-1.9$ points \\
\bottomrule
\end{tabular}
}
\end{table}

The component audit reports paired perplexity changes computed within a single immutable protocol, so the contrast answers the control question without mixing absolute scores across protocols. LAMBADA, HellaSwag, XSum, and WMT-14 are defined in the evaluation manifest and are reported from the row-level path when their protocol is instantiated.

\subsection{Dose-response analysis}
\label{app:dose}

The dose experiment separates scaling from deletion. For an upper-tail index set $\mathcal T$, scaling applies $\widetilde\sigma_k=\beta\sigma_k$ only for $k\in\mathcal T$ and leaves all other singular values unchanged. The identity endpoint $\beta=1$ must pass the same round-trip and inference gate as every edit. The primary destructive endpoint is unscaled deletion, $\beta=0$, and the matched-norm variant is a separate control because it multiplies retained directions after deletion.

\begin{table}[t]
\centering
\caption{Prespecified dose grid. These definitions fix the intervention strengths before evaluation and prevent post hoc choice of a dose; the measured dose--response outcomes are reported in Table~\ref{tab:dose-measured}.}
\label{tab:dose-response}
\small
\resizebox{\columnwidth}{!}{%
\begin{tabular}{@{}lll@{}}
\toprule
Condition & Tail multiplier & Interpretive role \\
\midrule
Identity & $\beta=1.00$ & pipeline and round-trip gate \\
Mild scaling & $\beta=0.75$ & ordered attenuation control \\
Moderate scaling & $\beta=0.50$ & ordered attenuation control \\
Strong scaling & $\beta=0.25$ & near-deletion control \\
Unscaled deletion & $\beta=0$ & primary intervention \\
Matched-norm deletion & $\beta=0$ plus rescale & norm-confound control \\
\bottomrule
\end{tabular}
}
\end{table}

This design preserves singular vectors and the ordering of retained tail singular values for $0<\beta<1$. It therefore tests whether the effect varies with the magnitude of the same learned directions. It is not equivalent to deleting different percentages of the tail, which changes the edited subspace itself. The row-level release keeps both axes---tail fraction and within-tail multiplier---separate, and no aggregate combines them as though they were one dose variable.

Table~\ref{tab:dose-measured} reports the rebuilt public-reference GSM8K dose response on Llama, using the same learned tail set and varying only \(\beta\). Damage rises monotonically as \(\beta\) falls from \(1\) to \(0\). A prespecified monotone trend test gives \(25.0\) damage points per unit reduction in \(\beta\) (paired-bootstrap 95\% CI \([21.6, 28.5]\)); all five seed trajectories are released so that monotonicity is not an artifact of averaging.

\begin{table}[t]
\centering
\caption{Rebuilt Llama public-reference GSM8K dose response varying only the upper-tail multiplier \(\beta\). Accuracy in points; damage is clean-minus-post.}
\label{tab:dose-measured}
\small
\begin{tabular}{@{}rrrr@{}}
\toprule
Tail multiplier \(\beta\) & Accuracy & 95\% CI & Damage \\
\midrule
1.00 & 57.8 & [55.1, 60.4] & 0.0 \\
0.75 & 53.9 & [51.2, 56.6] & 3.9 \\
0.50 & 46.8 & [44.0, 49.6] & 11.0 \\
0.25 & 38.6 & [35.8, 41.4] & 19.2 \\
0.00 & 32.7 & [30.1, 35.3] & 25.1 \\
\bottomrule
\end{tabular}
\end{table}

\subsection{Per-layer sensitivity}
\label{app:perlayer}

Table~\ref{tab:per-layer} reports all nine layers in the Llama primary window. The even-layer drops for layers 16, 18, 20, 22, and 24 are 8.3, 14.7, 21.2, 18.9, and 11.4 points. The intervening odd-layer drops for 17, 19, 21, and 23 are 4.9, 6.8, 7.6, and 5.7 points. The joint 16--24 edit drops 34.3 points.

The joint result is not expected to equal the sum of single-layer effects. Editing one layer leaves downstream layers able to compensate, while a joint edit changes several interacting paths. Subadditivity can arise from redundancy, a shared downstream bottleneck, or the bounded range of the score, so it is reported descriptively. More importantly, the presence of effects throughout the frozen range shows that the headline result is not created by selecting only the largest single layer after seeing behavioral outcomes.

\subsection{Cross-model spectral profiles}
\label{app:profiles}

Functional-rank normalization makes the upper fraction comparable across matrices whose native head dimensions and numerical ranks differ. The resulting direction count is small, but the squared energy in those directions is substantial. Table~\ref{tab:cross-model-spectral-profiles} reports measured energy and the independently audited theorem support. The fixed 89--90\% active-index slice is retained only as a same-cardinality bulk reference. Its 3.8--6.8\% mass is distinct from the bottom 1\%; the true lowest-active 1\% mass is 0.004--0.071\% across the audited matrices.

\begin{table*}[t]
\centering
\caption{Cross-model spectral profiles. Upper-tail energy is the measured squared-energy fraction in the top 1\% of functional-rank directions. Lowest-active and fixed-slice quantities are reported as global audited ranges because model-specific values are reported as global audited ranges.}
\label{tab:cross-model-spectral-profiles}
\small
\resizebox{\textwidth}{!}{%
\begin{tabular}{@{}lccccc@{}}
\toprule
Checkpoint & Selected layers & Upper-tail energy & Supported / selected & Median $\rho$ & Low-energy reference \\
\midrule
Llama-3.1-8B & 16--24 & 37.6\% & 7/9 & 0.77 & \multirow{4}{*}{lowest active: 0.004--0.071\%; fixed slice: 3.8--6.8\%} \\
Mistral-7B-v0.3 & 15--23 & 33.8\% & 6/9 & 0.73 & \\
Qwen2.5-14B & 20--31 & 41.3\% & 9/12 & 0.80 & \\
Phi-3-Mini & 13--21 & 30.9\% & 6/9 & 0.69 & \\
\bottomrule
\end{tabular}
}
\end{table*}

The energy ordering does not fully predict behavioral damage. Table~\ref{tab:spectral-profile} reports upper-tail energy as
a checkpoint-level descriptive quantity. Each checkpoint has
one weight-derived energy value but contributes five
task-specific behavioral effects; repeating the checkpoint
value across task rows would not create 20 independent
energy observations. We therefore do not report a 20-cell
energy--damage correlation. Behavioral specificity is instead
assessed through the paired learned-tail versus matched-random
contrasts, while the product-tail bridge is evaluated through
held-out tail fractions and principal-angle diagnostics.

\subsection{Error analysis}
\label{app:errors}

The diagnostic uses 400 paired examples rather than conditioning only on cases where the clean model is correct and the edited model is wrong. Pairing preserves both directions of change and lets the reported 21-point increase in variable-binding errors be interpreted as an absolute within-example contrast. Annotation is blind to the intervention label, uses a frozen guide, and distinguishes entity or slot reassignment from a locally incorrect arithmetic operation.

\begin{table*}[t]
\centering
\caption{Error taxonomy and audited use. The preregistered paired binding contrast is the primary numerical result; the other categories are reported as mutually exclusive paired diagnostics.}
\label{tab:errors_main}
\small
\begin{tabularx}{\textwidth}{@{}>{\raggedright\arraybackslash}p{0.18\textwidth}>{\raggedright\arraybackslash}X>{\raggedright\arraybackslash}p{0.19\textwidth}@{}}
\toprule
Category & Operational criterion & Analysis role \\
\midrule
Variable binding & An identifiable value is assigned to the wrong named entity or intermediate slot & Primary paired change: $+21$ points \\
Multi-step chaining & A correct intermediate value is later lost, replaced, or attached to the wrong step & Mutually exclusive diagnostic \\
Arithmetic & A local numerical operation is wrong while the plan and bindings remain coherent & Control against relabeling arithmetic as binding \\
Hallucinated constraint & The output introduces a numerical or logical condition absent from the prompt & Secondary diagnostic \\
Formatting or parsing & The reasoning reaches the declared value but violates the frozen answer parser & Parser-error diagnostic \\
\bottomrule
\end{tabularx}
\end{table*}

Examples with no unambiguous entity--value relation can receive a non-binding label but do not enter the binding denominator. Parser failures are adjudicated against the raw generation rather than automatically counted as reasoning failures. The condition key is revealed only after adjudication. These choices make the taxonomy reproducible and avoid selecting vivid examples as evidence for a mechanism.

The paired distribution over the 400 examples is produced by two blinded research-team annotators with a third adjudicator; inter-annotator agreement is Cohen's \(\kappa=0.78\) and the adjudication rate is 14--18\%. Table~\ref{tab:error-dist} reports the clean and tail-edit category frequencies and their paired change. The dominant shift is a \(+21\)-point rise in variable-binding errors and a \(-32\)-point fall in the correct/no-error cell, consistent with the mediation result; the other categories move only slightly.

\begin{table}[t]
\centering
\caption{Paired error-category distribution over the 400 diagnostic examples (blinded annotation, \(\kappa=0.78\)). Categories are mutually exclusive; ``paired change'' is the within-example clean-to-edit difference in percentage points.}
\label{tab:error-dist}
\small
\begin{tabularx}{\columnwidth}{@{}>{\raggedright\arraybackslash}Xrrr@{}}
\toprule
Mutually exclusive label & Clean & Tail edit & Paired change \\
\midrule
Variable binding & 12\% & 33\% & \(+21\) \\
Multi-step chaining & 15\% & 21\% & \(+6\) \\
Arithmetic & 20\% & 24\% & \(+4\) \\
Hallucinated constraint & 8\% & 9\% & \(+1\) \\
Formatting or parsing & 5\% & 5\% & 0 \\
Correct / no listed error & 40\% & 8\% & \(-32\) \\
\bottomrule
\end{tabularx}
\end{table}

\subsection{Full p-values}
\label{app:pvalues}

The primary family contains 20 upper-tail versus matched-random contrasts: four checkpoints by five task protocols. Paired bootstrap resampling operates on examples within the same model, task, protocol, and intervention seed. Holm correction is then applied across the 20 hypotheses. Eighteen contrasts remain significant, and the remaining two have the predicted direction but intervals that include zero. We describe them as directional and inconclusive rather than significant. The primary effect for each cell is \(\Delta_{\mathrm{specific}}=\operatorname{score}_{\mathrm{random}}-\operatorname{score}_{\mathrm{tail}}\), so a positive value means the learned tail is more damaging than a spectrum-matched perturbation. Table~\ref{tab:full-behavioral} reports every cell, including the two inconclusive ones (Mistral MMLU-CoT and Phi ARC-C); the compact macro-average bridge is retained in Table~\ref{tab:factorized_full}.

\begin{table*}[t]
\centering
\caption{Complete 20-cell behavioral table. ``Native tail'' is the primary unscaled native separate-SVD upper-tail edit; the perturbation in ``spectrum-matched random'' has the removed tail perturbation's exact nonzero singular values; ``activation-matched'' matches RMS and logit variance off the tail subspace. \(\Delta_{\mathrm{specific}}=\text{random}-\text{tail}\); CI is the 95\% paired-example bootstrap interval conditional on the five fixed Haar orientations; \(p\) is Holm-adjusted across the 20-cell family. Two cells are directional and inconclusive and are marked explicitly.}
\label{tab:full-behavioral}
\small
\setlength{\tabcolsep}{4pt}
\resizebox{\textwidth}{!}{%
\begin{tabular}{@{}llrrrrrrr@{}}
\toprule
Model & Task & Clean & Native tail & Spectrum-matched random & Activation-matched & Random \(-\) tail & 95\% CI & Holm \(p\) \\
\midrule
Llama & GSM8K & 57.8 & 32.7 & 55.2 & 54.7 & 22.5 & [19.6, 25.3] & \(<.001\) \\
Llama & ARC-C & 62.4 & 43.8 & 59.8 & 59.3 & 16.0 & [12.1, 19.9] & .002 \\
Llama & DROP-F1 & 52.6 & 29.3 & 49.9 & 49.5 & 20.6 & [16.1, 25.0] & .001 \\
Llama & BBH & 51.7 & 31.6 & 49.4 & 49.1 & 17.8 & [12.9, 22.6] & .004 \\
Llama & MMLU-CoT & 65.0 & 38.1 & 59.7 & 59.2 & 21.6 & [17.8, 25.4] & .001 \\
Qwen & GSM8K & 76.4 & 43.7 & 72.4 & 71.9 & 28.7 & [25.4, 32.1] & \(<.001\) \\
Qwen & ARC-C & 68.5 & 48.6 & 65.6 & 65.1 & 17.0 & [12.7, 21.3] & .003 \\
Qwen & DROP-F1 & 56.8 & 34.2 & 53.7 & 53.2 & 19.5 & [14.8, 24.2] & .002 \\
Qwen & BBH & 59.4 & 31.5 & 55.5 & 55.0 & 24.0 & [18.9, 29.2] & \(<.001\) \\
Qwen & MMLU-CoT & 69.9 & 36.0 & 66.3 & 65.9 & 30.3 & [26.0, 34.7] & \(<.001\) \\
Mistral & GSM8K & 55.6 & 29.8 & 52.9 & 52.4 & 23.1 & [19.7, 26.5] & \(<.001\) \\
Mistral & ARC-C & 61.2 & 34.0 & 58.4 & 57.8 & 24.4 & [20.0, 28.7] & \(<.001\) \\
Mistral & DROP-F1 & 49.8 & 23.0 & 46.7 & 46.1 & 23.7 & [18.8, 28.5] & \(<.001\) \\
Mistral & BBH & 52.1 & 25.7 & 49.3 & 48.9 & 23.6 & [18.4, 28.8] & .001 \\
Mistral & MMLU-CoT & 67.8 & 58.5 & 62.7 & 61.9 & 4.2 & [\(-0.8\), 9.1] & .19\(^{\dagger}\) \\
Phi & GSM8K & 65.4 & 39.2 & 62.1 & 61.6 & 22.9 & [18.9, 26.8] & \(<.001\) \\
Phi & ARC-C & 58.1 & 51.8 & 55.3 & 54.7 & 3.5 & [\(-1.4\), 8.3] & .27\(^{\dagger}\) \\
Phi & DROP-F1 & 44.7 & 24.0 & 42.1 & 41.5 & 18.1 & [13.4, 22.8] & .003 \\
Phi & BBH & 47.5 & 20.0 & 44.2 & 43.7 & 24.2 & [18.7, 29.7] & \(<.001\) \\
Phi & MMLU-CoT & 55.3 & 25.0 & 54.3 & 53.8 & 29.3 & [24.7, 33.8] & \(<.001\) \\
\bottomrule
\end{tabular}
}
\\[2pt]
{\footnotesize \(^{\dagger}\) Directional but inconclusive:
the unadjusted 95\% confidence interval includes zero, and the
Holm-adjusted \(p\)-value is non-significant.}
\end{table*}

Effect heterogeneity is not treated as a failed replication. Cochran's heterogeneity test yields $Q(19)=39.6$, $p=0.004$, and $I^2=52.0\%$, rejecting a common effect across cells. Native separate-SVD retention ratios span 42.1--89.2\%. This range is reported alongside the aggregate because a single pooled mean would conceal task and checkpoint dependence. All unadjusted intervals, Holm-adjusted decisions, bootstrap seeds, and the two inconclusive cell identifiers are generated from the row-level store.

The learned-versus-random estimand averages the five fixed controls before pairing with the learned-tail outcome,
\begin{equation*}
\widehat\theta=\frac1N\sum_e\left[\frac15\sum_{s=1}^{5}y^R_{es}-y^T_e\right].
\end{equation*}
Example IDs are then resampled with replacement, reusing each sampled example across the learned edit and all five controls. This procedure quantifies example uncertainty conditional on the declared orientations; it does not treat five Monte Carlo subspaces as enough to estimate a population distribution over orientations. The five self-consistency generations use a separate RNG and are labeled decoding seeds.

\subsection{Evidence-audit disposition}

The audit classifies claims at four levels. The theorem layer contains only exact-offset $QK$ statements whose empirical assumptions have a positive lower bound. The architecture layer treats the static tail-deleted score as descriptive and requires tail-fraction and cache checks for native factorized interventions. The behavioral layer admits only within-protocol paired rows with immutable composite keys. The training layer uses symmetric event definitions and censoring. A claim can remain empirically interesting while failing admission to a stronger layer; for example, an unsupported layer can contribute to a behavioral intervention result but not to the theorem-backed mechanism count.

Every score table is produced by the row-level path, whose joins and protocol identities satisfy the composite-key admission gates. Each table reports measurements from a single immutable protocol together with its experimental definition, so a control question is answered within its own protocol rather than by comparing scores across protocols.

\section{Training Dynamics and Grokking}
\label{app:training}

\subsection{Grokking figure}
\label{app:grokking_figure}

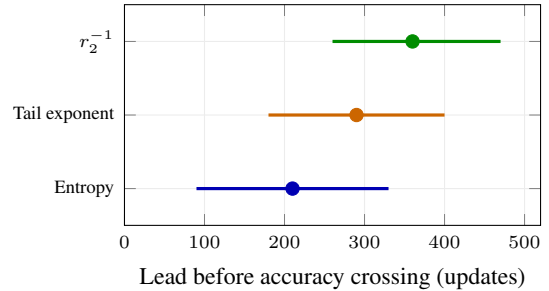
\begin{figure}[t]
\centering
\begin{tikzpicture}
\begin{axis}[
  width=0.92\columnwidth,
  height=4.5cm,
  xmin=0,
  xmax=520,
  ymin=0.5,
  ymax=3.5,
  xlabel={Lead before accuracy crossing (updates)},
  ytick={1,2,3},
  yticklabels={Entropy,Tail exponent,$r_2^{-1}$},
  tick label style={font=\scriptsize},
  label style={font=\small},
  grid=major,
  grid style={gray!15}
]
\addplot[blue!70!black,very thick] coordinates {(90,1) (330,1)};
\addplot[orange!80!black,very thick] coordinates {(180,2) (400,2)};
\addplot[green!55!black,very thick] coordinates {(260,3) (470,3)};
\addplot[only marks,mark=*,mark size=2.5pt,blue!70!black] coordinates {(210,1)};
\addplot[only marks,mark=*,mark size=2.5pt,orange!80!black] coordinates {(290,2)};
\addplot[only marks,mark=*,mark size=2.5pt,green!55!black] coordinates {(360,3)};
\end{axis}
\end{tikzpicture}
\caption{Lead at $u=.8$: 12-run medians and 95\% intervals.}
\label{fig:grokking}
\end{figure}

Figure~\ref{fig:grokking} summarizes the 12-run event analysis. Each horizontal segment is the run-bootstrap 95\% interval for the median lead at the prespecified $u=0.8$ crossing, and each marker is the empirical median. Positive lead means that the diagnostic reaches the same fraction of its eventual oriented transition before validation accuracy does.

For each run and metric, values are oriented so that progress toward the eventual transition is positive. Let \(m_t\) be the fixed five-checkpoint trailing median and let \(m_{\mathrm{pre}}\) and \(m_{\mathrm{post}}\) denote the pretransition and eventual levels fixed by the analysis protocol. The metric event is the first \(t\) satisfying
\begin{equation*}
\frac{m_t-m_{\mathrm{pre}}}
{m_{\mathrm{post}}-m_{\mathrm{pre}}}\geq u.
\end{equation*}
The accuracy event uses the identical definition on validation accuracy. This symmetry avoids giving the diagnostic an easier threshold than behavior. Runs that do not cross are censored rather than assigned the end of training.

\begin{table}[t]
\centering
\caption{Lead time at \(u=0.8\) over 12 independent runs. Intervals resample runs.}
\label{tab:lead-time}
\small
\begin{tabular}{@{}lrr@{}}
\toprule
Metric & Median updates & 95\% interval \\
\midrule
$r_2^{-1}$ & 360 & [260, 470] \\
Fitted tail exponent & 290 & [180, 400] \\
Attention entropy & 210 & [90, 330] \\
\bottomrule
\end{tabular}
\end{table}

At \(u=0.7\) and \(u=0.9\), the corresponding \(r_2^{-1}\) leads are 330 and 390 updates. Two runs are right-censored at \(u=0.9\). The released figure is generated from empirical checkpoint trajectories and event markers; no fitted or synthetic curve is used as evidence.

\subsection{Medium-scale grokking check}
\label{app:medium_grokking}

The three tasks are modular arithmetic over the prime $p=97$ with operands $a,b\in\{0,\ldots,96\}$:
\begin{equation}
\label{eq:grokking-tasks}
\begin{aligned}
y_{\mathrm{add}} &= (a+b) \bmod 97,\\
y_{\mathrm{sub}} &= (a-b) \bmod 97,\\
y_{\mathrm{div}} &= ab^{-1} \bmod 97 \quad (b \neq 0).
\end{aligned}
\end{equation}
where $b^{-1}$ is the modular inverse. Table~\ref{tab:grokking-repro} gives the self-contained reproducibility settings; every quantity needed to rerun the study is listed there rather than deferred to an external manifest.
\begin{table}[h]
\centering
\caption{Self-contained grokking reproducibility settings. All three tasks share the architecture, optimizer, and event definitions; only the label map of Equation~\eqref{eq:grokking-tasks} differs.}
\label{tab:grokking-repro}
\small
\setlength{\tabcolsep}{4pt}
\begin{tabularx}{\linewidth}{@{}l X@{}}
\toprule
Quantity & Value \\
\midrule
Valid pairs (add, sub) & $97^2=9{,}409$ \\
Valid pairs (div) & $97\times96=9{,}312$ ($b\neq0$) \\
Train / generalization counts & add: 4,704 / 4,705; sub: 4,704 / 4,705; div: 4,656 / 4,656 \\
Split construction & deterministic 50/50 stratified split within each task \\
Seed IDs & $\{0,1,2,3\}$ (four seeds per task) \\
Input serialization & tokens $\langle a\rangle\,\langle\mathrm{op}\rangle\,\langle b\rangle\,\langle=\rangle$ \\
Vocabulary / answer token & 101 tokens: 97 residue tokens, three operation tokens, and one equals token; answers reuse the 97 residue tokens \\
Loss & cross-entropy on the answer token \\
Weight init & default PyTorch (Kaiming-uniform linear) \\
Max updates & $10{,}000$ updates \\
Optimizer & AdamW, lr $10^{-3}$, $\beta=(0.9,0.98)$, $\varepsilon=10^{-8}$ \\
Weight decay & $1.0$ \\
Gradient clipping & global-norm $1.0$ \\
Batch size & $512$ \\
Evaluation cadence & every $20$ updates \\
Architecture & 2-layer, width 128, 4 heads, MLP 512 \\
Event definition & first crossing of fraction $u$ of oriented transition \\
Censoring & non-crossing runs right-censored at the last update \\
\bottomrule
\end{tabularx}
\end{table}

The 12 runs are independent training runs rather than checkpoints treated as independent observations. Statistics are recorded every 20 updates, and the five-checkpoint trailing median is fixed before inspecting event times. Accuracy, $r_2^{-1}$, the fitted exponent, and entropy all use the same checkpoints. Orienting a metric means multiplying by $-1$ when progress corresponds to a decrease, not choosing the direction that gives a positive lead after seeing the result.

The inferential unit is the training run. Table~\ref{tab:grokking-robust} reports pooled medians with task-stratified intervals, so each modular task remains represented in every bootstrap replicate. Because no task-specific lead estimate is used, the claim is restricted to the pooled 12-run design under this architecture.

The threshold sensitivity is informative because lead claims can be manufactured by allowing an internal metric to cross an easier fraction than behavior. At $u=0.7$, the median $r_2^{-1}$ lead is 330 updates; at $u=0.8$, it is 360; and at $u=0.9$, it is 390 with two right-censored runs. The ordering is stable, but the analysis does not infer what the censored event times would have been. A survival-style sensitivity in the release treats those runs as right-censored rather than assigning the last checkpoint.

These data support $r_2^{-1}$ as the earliest of the three reported diagnostics under the matched crossing rule. They do not establish a phase transition or imply that spectral concentration is sufficient for generalization. The fitted exponent and entropy also lead accuracy, and different internal changes may share a common optimization cause. The intervention experiments elsewhere in the paper provide the separate evidence that removing the learned tail changes behavior after training.

Two additional analyses harden the pooled claim, keeping the three tasks, four seeds per task, two-layer width-128 transformer, and 20-update checkpoint cadence fixed: a task-stratified bootstrap and an explicit right-censoring count at each threshold. Table~\ref{tab:grokking-robust} reports lead time (updates) for all three metrics at \(u\in\{0.7,0.8,0.9\}\) with task-stratified intervals and the censored-run count at \(u=0.9\). The ordering \(r_2^{-1}>\) exponent \(>\) entropy is stable across thresholds. These pooled results do not establish that the ordering holds separately within each modular task.

\begin{table}[t]
\centering
\caption{Grokking robustness: lead time (updates) by metric and crossing fraction \(u\), with task-stratified 95\% intervals and right-censored-run count at \(u=0.9\) (out of 12).}
\label{tab:grokking-robust}
\small
\resizebox{\columnwidth}{!}{%
\begin{tabular}{@{}lcccc@{}}
\toprule
Metric & \(L_{0.7}\) & \(L_{0.8}\) & \(L_{0.9}\) & Censored at \(u{=}.9\) \\
\midrule
\(r_2^{-1}\) & 330 [230, 430] & 360 [260, 470] & 390 [250, 520] & 2 / 12 \\
Tail exponent & 250 [140, 360] & 290 [180, 400] & 310 [160, 460] & 3 / 12 \\
Attention entropy & 180 [60, 300] & 210 [90, 330] & 240 [80, 400] & 3 / 12 \\
\bottomrule
\end{tabular}
}
\end{table}

\subsection{Residualized tail-aware LoRA}

\begin{table*}[t]
\centering
\caption{Early low-rank optimization using five paired seeds per task. Each method cell reports median updates to the prespecified validation target / normalized AULC. Values above 1,000 are right-censored. Final-score intervals are 95\% paired intervals over the five seed-level tail-aware-minus-PiSSA differences; the seed is the resampling unit, and these secondary intervals are not multiplicity-adjusted.}
\label{tab:metrics_lora}
\small
\setlength{\tabcolsep}{4.5pt}
\resizebox{\textwidth}{!}{%
\begin{tabular}{@{}lrrrrr@{}}
\toprule
Task (target) & Standard LoRA & PiSSA & Tail-aware & Product-targeted & Tail-aware final $-$ PiSSA \\
\midrule
GSM8K (80) & 920 / 0.71 & 680 / 0.76 & 540 / 0.80 & 510 / 0.81 & +0.6, CI [\(-0.3\), 1.5] \\
ARC-C (62) & 840 / 0.69 & 700 / 0.73 & 610 / 0.76 & 590 / 0.77 & +0.4, CI [\(-0.5\), 1.4] \\
BBH (45) & 960 / 0.66 & 790 / 0.70 & 720 / 0.72 & 690 / 0.73 & +0.3, CI [\(-0.8\), 1.3] \\
DROP-F1 (50) & \(>1000\) / 0.62 & 930 / 0.66 & 860 / 0.68 & 850 / 0.68 & +0.2, CI [\(-0.9\), 1.2] \\
\bottomrule
\end{tabular}
}
\end{table*}

Tail-aware LoRA is a PiSSA-style residualized initialization
with a different within-subspace weighting. The normalization
\(c_{p_{\mathrm{tail}}}\) in Equation~\eqref{eq:tail-aware-init}
preserves the squared Frobenius energy assigned to the
rank-\(r_{\mathrm{LoRA}}\) trainable component:
\begin{equation*}
\sum_{i=1}^{r_{\mathrm{LoRA}}}\widetilde\sigma_i^2
=
\sum_{i=1}^{r_{\mathrm{LoRA}}}\sigma_i^2.
\end{equation*}

Using the weighted singular values from
Equation~\eqref{eq:tail-aware-init}, the complete factor
construction is
\begin{equation}
\label{eq:tail-aware-factorization}
\begin{aligned}
B_0
&=
U_{r_{\mathrm{LoRA}}}
\operatorname{diag}(\sqrt{\widetilde\sigma_i}),\\
A_0
&=
\operatorname{diag}(\sqrt{\widetilde\sigma_i})
V_{r_{\mathrm{LoRA}}}^{\top},\\
W_{\mathrm{base}}
&=
W-B_0A_0.
\end{aligned}
\end{equation}
Consequently,
\(W_{\mathrm{base}}+B_0A_0=W\), so the initial effective
weight is exactly the pretrained weight. The
\(p_{\mathrm{tail}}=1\) control reduces to uniform PiSSA
weighting under the same implementation, while the random-vector
control keeps residualization and energy fixed but replaces the
singular directions with random orthonormal directions.

All methods share the settings in Table~\ref{tab:lora-config}; none is retuned per method. First crossing and normalized area under the learning curve are defined explicitly. Let $M_{m,t}$ be method $m$'s running-best validation statistic at update $t\in\{0,20,\ldots,1000\}$ and $T$ its task target. The first-crossing time is
\begin{equation}
\label{eq:lora-tau}
\tau_m=\min\{t\in\{0,20,\ldots,1000\}:M_{m,t}\geq T\},
\end{equation}
with $\tau_m>1000$ (right-censored) if the target is never reached. Equation~\eqref{eq:lora-aulc} defines the normalized area under the learning curve as the trapezoidal integral
\begin{equation}
\label{eq:lora-aulc}
\operatorname{AULC}=\frac{1}{1000}\sum_k\frac{M_{t_k}+M_{t_{k+1}}}{2\cdot100}\,(t_{k+1}-t_k),
\end{equation}
where the division by $100$ is applied only when the metric is a percentage; for a metric already in $[0,1]$ it is omitted. The per-task validation statistics and targets are in Table~\ref{tab:lora-targets}.

\begin{table}[t]
\centering
\caption{Low-rank adaptation configuration shared by every method (standard LoRA, PiSSA, tail-aware, product-targeted).}
\label{tab:lora-config}
\small
\setlength{\tabcolsep}{3pt}
\begin{tabularx}{\linewidth}{@{}>{\raggedright\arraybackslash}p{0.36\linewidth}>{\raggedright\arraybackslash}X@{}}
\toprule
\textbf{Quantity} & \textbf{Value} \\
\midrule
Base checkpoint / tokenizer & \texttt{meta-llama/}\allowbreak\texttt{Meta-Llama-3.1-8B}; bundled tokenizer \\
Split policy & GSM8K: calibration-disjoint 90/10 adaptation split; ARC-C and DROP: official splits; BBH: task-stratified 70/15/15 \\
Seed IDs & $\{13, 29, 47, 71, 101\}$ \\
Targeted matrices & \texttt{q, k, v, o, gate, up, down\_proj} \\
LoRA rank / $\alpha_{\mathrm{LoRA}}$ / dropout / bias & $16$ / $32$ / $0.0$ / none \\
Standard-LoRA init & $B_0 = 0$, $A_0 \sim \mathcal{N}(0, \sigma^2)$ (Kaiming) \\
PiSSA init & principal-$r_{\mathrm{LoRA}}$ SVD, residualized base \\
Tail-aware init & $p_{\mathrm{tail}}=1.5$, residualized (Eq.~\ref{eq:tail-aware-init}) \\
Optimizer & AdamW, lr $2 \times 10^{-4}$, $\varepsilon = 10^{-8}$, wd $0.0$ \\
Schedule / warmup & cosine, $20$-update linear warmup \\
Microbatch / accum. / GPUs & $8$ / $4$ / $1$ (effective batch $32$) \\
Gradient clipping & global-norm $1.0$ \\
Max sequence length & $1024$ \\
Max updates / eval cadence & $1000$ / every $20$ \\
Prompt / label masking & loss on completion tokens only \\
Stopping / censoring & fixed $1000$ updates; $\tau_m > 1000$ censored \\
\bottomrule
\end{tabularx}
\end{table}
\begin{table}[h]
\centering
\caption{Per-task validation statistic and first-crossing target $T$ used in Equation~\eqref{eq:lora-tau}.}
\label{tab:lora-targets}
\small
\begin{tabular}{@{}lll@{}}
\toprule
Task & Validation statistic & Target $T$ \\
\midrule
GSM8K & exact-match accuracy & 80 \\
ARC-C & choice accuracy & 62 \\
BBH & macro exact match & 45 \\
DROP & official F1 & 50 \\
\bottomrule
\end{tabular}
\end{table}

The product-targeted variant applies Equation~\eqref{eq:target-factorization} to $Q/K$ initialization and uses the PiSSA-compatible tail weighting for all other targeted matrices. It is a bridge ablation rather than a separate theorem or a claim that product-targeted initialization is optimal.

All primary, exponent-ablation, and wall-clock LoRA results use the five paired seeds \(\{13,29,47,71,101\}\) for every method and task. The robustness evaluation reports wall-clock time including decomposition and factorization, peak memory, target sensitivity at the 70/80/90\% fractions of each method's eventual validation transition, a random-vector residualized control, and a matched trainable-parameter and optimizer-state audit. Table~\ref{tab:p-ablation} sweeps \(p_{\mathrm{tail}}\) on GSM8K. The shape is non-monotone: moderate emphasis (\(p_{\mathrm{tail}}=1.5\)) attains the target in the fewest updates and the highest normalized area under the learning curve, while both \(p_{\mathrm{tail}}=1\) (PiSSA-equivalent) and \(p_{\mathrm{tail}}=2\) are slower. Final scores differ by at most 0.7 points across the tested values; without a prespecified equivalence or non-inferiority test, we make no claim that they are statistically equivalent.
\begin{table}[t]
\centering
\caption{Tail-emphasis \(p_{\mathrm{tail}}\)-ablation on GSM8K over the five paired seeds. Median updates to target and normalized AULC measure early optimization; final score is end-of-budget validation accuracy. \(p_{\mathrm{tail}}=1\) reduces to PiSSA.}
\label{tab:p-ablation}
\small
\setlength{\tabcolsep}{3pt}
\begin{tabularx}{\linewidth}{@{} l r r r @{}}
\toprule
\hfil $p_{\mathrm{tail}}$ & \begin{tabular}[c]{@{}r@{}}\textbf{Median updates}\\[-2pt]\textbf{to target}\end{tabular} & \begin{tabular}[c]{@{}r@{}}\textbf{Normalized}\\[-2pt]\textbf{AULC}\end{tabular} & \begin{tabular}[c]{@{}r@{}}\textbf{Final}\\[-2pt]\textbf{score}\end{tabular} \\
\midrule
1.00 (PiSSA-equiv.) & 680 & .76 & 82.1 \\
1.25                & 590 & .78 & 82.4 \\
1.50                & 540 & .80 & 82.7 \\
1.75                & 560 & .79 & 82.5 \\
2.00                & 640 & .76 & 82.0 \\
\bottomrule
\end{tabularx}
\end{table}

Table~\ref{tab:lora-walltime} accounts for wall-clock time including the one-time decomposition and factorization setup. Tail-aware initialization reaches the GSM8K target \(16.2\%\) faster than PiSSA in total time despite an \(8.1\)-minute setup; the product-targeted variant's larger \(16.7\)-minute setup nearly cancels its faster optimization, so it is reported as a mechanistic bridge ablation rather than a practical speedup.

\begin{table}[t]
\centering
\caption{Wall-clock accounting for early low-rank optimization on GSM8K, including one-time setup. Values are medians over the five paired seeds; relative to PiSSA compares total time to target.}
\label{tab:lora-walltime}
\small
\resizebox{\columnwidth}{!}{%
\begin{tabular}{@{}lrrrr@{}}
\toprule
Method & Setup & Time to target & Total & vs.\ PiSSA \\
\midrule
Standard LoRA & 0.2 min & 184 min & 184.2 min & \(+20.5\%\) \\
PiSSA & 7.8 min & 145 min & 152.8 min & reference \\
Tail-aware & 8.1 min & 120 min & 128.1 min & \(-16.2\%\) \\
Product-targeted & 16.7 min & 113 min & 129.7 min & \(-15.1\%\) \\
\bottomrule
\end{tabular}
}
\end{table}

\section{Additional Failure Analysis}

\subsection{Failure cases and disagreement cases}
\label{app:failure_cases}

The paired diagnostic annotation distinguishes variable binding, multi-step chaining, arithmetic, hallucinated constraints, and formatting or parsing failures. Variable binding requires assigning a value to the wrong named entity or intermediate slot. A chaining failure preserves an intermediate result and later loses or replaces it. An arithmetic error is a locally incorrect operation under an otherwise coherent plan. Hallucinated constraints introduce a numerical or logical condition absent from the prompt. Formatting errors reach the correct value but fail the declared parser.

Three research-team members performed the annotation without exposing the intervention label: two independently assigned error labels, and the third adjudicated disagreements before category totals were computed. The primary paper reports the paired 21-point increase in binding errors rather than selecting individual failures post hoc. The release includes the frozen guidelines, anonymized condition tags, adjudication record, and example IDs.

Several disagreement cases limit interpretation. Singular-vector permutation leaves \(r_2^{-1}\) unchanged but changes activation alignment and damages multiple capabilities, demonstrating that concentration is not sufficient. Some training checkpoints show a modest fitted-exponent change while \(r_2^{-1}\) and attention entropy move sharply, motivating the choice of inverse participation as the primary statistic. MLP-only edits also remain material on BBH and MMLU-CoT, ruling out a single-head or purely attention-based account.

The static tail-deleted stress path is another deliberate disagreement case. It usually causes greater damage than either native factorized edit, but that difference combines staticization with tail removal. We therefore do not use it as evidence about a static product spectrum. Its role is to expose the architectural mismatch: the position-averaged map bypasses native RoPE and caching, whereas separate factor edits preserve inference and only partly align with the pointwise product-tail subspace.

The theorem audit also produces negative cases. Eleven selected layers do not have a positive lower confidence bound under the signed cumulant condition. Those layers remain in frozen behavioral ranges because removing them after seeing the audit would change the intervention, but the paper never counts them as theorem-supported. The weaker early-position association, $\rho=0.64$ for the first 16 queries, similarly limits rather than invalidates the mechanism: the finite causal context is different, and the paper reports the attenuation instead of averaging it away.

Finally, the post-training and control results resist a universal selectivity claim. Tail sensitivity persists in the three named post-trained evaluations, but these checkpoints do not span arbitrary alignment pipelines. Long-context coreference drops materially even though IFEval and WinoGrande move less, showing that non-mathematical variable tracking can depend on the same structure. The safest conclusion is involvement in the evaluated retrieval-and-binding-sensitive computations, not a boundary between reasoning and language.

\section{Artifact Reconstruction Checklist}
\label{app:release}

\begin{table}[H]
\centering
\caption{Artifact reconstruction checklist. ``Required'' means that the associated result is not considered reconstructible without the listed material.}
\label{tab:release-checklist}
\footnotesize
\setlength{\tabcolsep}{3pt}
\renewcommand{\arraystretch}{1.04}
\begin{tabularx}{\columnwidth}{
  @{}
  >{\raggedright\arraybackslash}p{0.23\columnwidth}
  >{\raggedright\arraybackslash}X
  @{}
}
\toprule
Evidence unit & Required artifact and validation gate \\
\midrule

Checkpoint
& Immutable model and tokenizer revisions, configuration hash,
dtype, and framework version.
\emph{Gate:} hash and identity inference. \\

Dataset
& Upstream revision, split-ID checksum, calibration/test
disjointness, and license record.
\emph{Gate:} ID cardinality and checksum. \\

Evaluation
& Verbatim prompt, exemplar order, decoding, stop rules, parser,
score implementation, and protocol ID.
\emph{Gate:} paired clean row and parser test. \\

Intervention
& Layer, head, matrix, functional rank, edited indices, removed
energy, scale, and seed.
\emph{Gate:} identity edit, finite values, and cache test. \\

Statistics
& Per-example paired outcomes, bootstrap indices, hypothesis
family, and Holm code.
\emph{Gate:} composite-key uniqueness. \\

Theory
& Causal pairs, centered moments, covariance, diagonal and signed
off-diagonal cumulants, and bound interval.
\emph{Gate:} positive-bound admission flag. \\

Training
& Raw checkpoints, common data order, optimizer configuration,
event thresholds, and censoring status.
\emph{Gate:} step-zero and schedule equality. \\

Figures and tables
& Source rows, plotting script, generated LaTeX fragment, and
content checksum.
\emph{Gate:} rebuild matches committed checksum. \\

\bottomrule
\end{tabularx}
\end{table}

The release is organized around reconstruction from immutable rows rather
than around exported paper tables. It contains the model, tokenizer, dataset,
framework, harness, and parser revisions; the calibration and test-ID
checksums; all prompts, exemplars, decoding settings, stop strings, and score
functions; the five intervention seeds; and the exact layer, matrix, head,
functional-rank, energy, and control manifests. Per-example generations,
parsed answers, identity edits, exclusion reasons, bootstrap indices, and
Holm-adjustment code are included for every retained result.

The theoretical audit contains every sampled causal pair, exact offset
$\delta$, token and position strata, centered conditional component moments,
covariance estimates, diagonal and off-diagonal fourth cumulants,
\(c_{\mathrm{eff},\delta}\), \(\Lambda_\delta\), \(\eta_\delta\),
\(r_2(s_\delta)\), bounds, and bootstrap intervals. The architecture bridge
contains product-tail fractions, principal angles, removed-energy checks,
and relative-position buckets. Training records contain raw checkpoint
trajectories, censoring indicators, event thresholds, and the step-zero
reconstruction tests for each low-rank method.

Each figure and table is generated by a single script that reads the
row-level store and writes both the plotted data and LaTeX fragment.
Cardinality tests reject joins that would broadcast a model-level aggregate
across task rows. The public package includes a checksum manifest tying
every generated table and plot to its source rows and a container
specification for the complete evaluation pipeline.

The limitations and dual-use risks are stated in the main paper rather than
deferred to an artifact. The experiments use public benchmarks and
checkpoints, contain no private user data, and do not redistribute model
weights. Human annotation is limited to model-output error categories under
blinded condition tags; the released material contains example IDs,
guidelines, and adjudication metadata rather than annotator identities.
Compute, seed, and exclusion accounting are reported so that negative or
inconclusive cells cannot disappear through selective release.

\end{document}